\documentclass[11pt]{article}

\PassOptionsToPackage{table}{xcolor}  

\usepackage[final]{acl}
\usepackage{booktabs}
\usepackage[table]{xcolor}  

\usepackage{arydshln} 
\usepackage{graphicx}
\usepackage{amssymb} 
\usepackage[most]{tcolorbox}  
\usepackage{enumitem}
\usepackage{siunitx}
\usepackage{times}
\usepackage{latexsym}
\usepackage{multirow}
\usepackage[T1]{fontenc}
\usepackage{longtable}

\usepackage[normalem]{ulem} 

\usepackage[utf8]{inputenc}
\usepackage{booktabs}
\usepackage{microtype}

\usepackage{inconsolata}

\usepackage{subcaption}

\usepackage{color,soul}
\usepackage{array}
\newcolumntype{H}{>{\setbox0=\hbox\bgroup}c<{\egroup}@{}}

\newcommand{\mee}[1]{\textcolor{red}{AA: #1}}

\definecolor{mycustomcolor}{RGB}{74, 0, 255}

\title{Explain-then-Process: Using Grammar Prompting to Enhance Grammatical Acceptability Judgments}

\author{
  Russell Scheinberg\textsuperscript{1} \quad
  Ameeta Agrawal\textsuperscript{1} \quad
  Amber Shore\textsuperscript{1} \quad
  So Young Lee\textsuperscript{2} \\[4pt]
  \textsuperscript{1}\,Portland State University, USA \\
  \textsuperscript{2}\,Miami University, USA \\[4pt]
  \texttt{\{rschein2, ameeta, ashore\}@pdx.edu} \quad
  \texttt{soyoung.lee@miamioh.edu}
}

\begin{document}
\maketitle
\begin{abstract}

Large language models (LLMs) can explain grammatical rules, yet they often fail to apply those rules when judging sentence acceptability. We present \emph{grammar prompting}, an explain-then-process paradigm: a large LLM first produces a concise explanation of the relevant syntactic phenomenon, then that explanation is fed back as additional context to the \emph{target model} -- either an LLM or a smaller language model (SLM) -- before deciding which sentence of a minimal pair is grammatical. On the English BLiMP, Chinese SLING, and Russian RuBLiMP benchmarks, this simple prompt design yields substantial improvements over strong baselines across a wide range of syntactic phenomena. 
Feeding an LLM’s metalinguistic explanation back to the target model bridges the gap between \emph{knowing} a rule and \emph{using} it. On SLMs, grammar prompting alone trims the average LLM-SLM accuracy gap by 20\%, and when paired with chain-of-thought, by 56\% (13.0 pp $\to$ 5.8 pp), all at negligible cost.
The lightweight, language-agnostic cue lets low-cost SLMs approach frontier-LLM performance in multilingual settings.




\end{abstract}

\section{Introduction}

Large Language Models (LLMs) occupy an uncanny valley: they appear \textit{almost} human in their language use, excelling at tasks from text generation to translation to complex reasoning. Yet these same models show surprising gaps, particularly when asked to make explicit grammatical judgments - a limitation that becomes especially apparent in multilingual contexts \cite{alrajhi-etal-2022-assessing, suijkerbuijkblimp}. While {modern language models} generate fluent text, they often struggle to reflect on grammatical structure, suggesting a disconnect between their ability to use language and their capacity to reason about it{, a gap that is particularly pronounced in smaller models}.
 
Consider this example. We asked Claude 3.5 Sonnet to reason out which minimal pair sentence \cite{warstadt-etal-2020-blimp-benchmark} is more grammatically acceptable:

\begin{tcolorbox}[
  colframe=black, colback=purple!10, 
  boxrule=0.8pt, sharp corners, 
  width=\columnwidth, 
]
\begin{itemize}[leftmargin=1em, nosep]
  \item[(A)] Only a popsicle that Danielle admires ever freezes. \checkmark
  \item[(B)] A popsicle that only Danielle admires ever freezes. $\times$
\end{itemize}
\end{tcolorbox}

Sonnet incorrectly selected `\textbf{B}'. Its reasoning has three steps: \textbf{i}, it paraphrases the meanings of the sentences as (A), ``The only thing that ever freezes is a popsicle Danielle admires'' and (B), ``A popsicle freezes only if Danielle, and no one else, admires it''; \textbf{ii}, Sonnet {analyzes the paraphrases}, noting that both (A) and (B) have the same ``only'' + ``ever'' pattern; and \textbf{iii}, Sonnet {concludes} that ``both are grammatical'', and chooses (B) as `clearer'. We speculate that the  paraphrase hides the fact that ``ever'' is a negative-polarity item (NPI) which must be within the scope of ``only.'' In (A), ``only'' properly licenses ``ever'', but in (B), it does not. By reducing both sentences to paraphrases, the model loses track of this licensing constraint and declares (B) acceptable.

\begin{figure*}[!t]
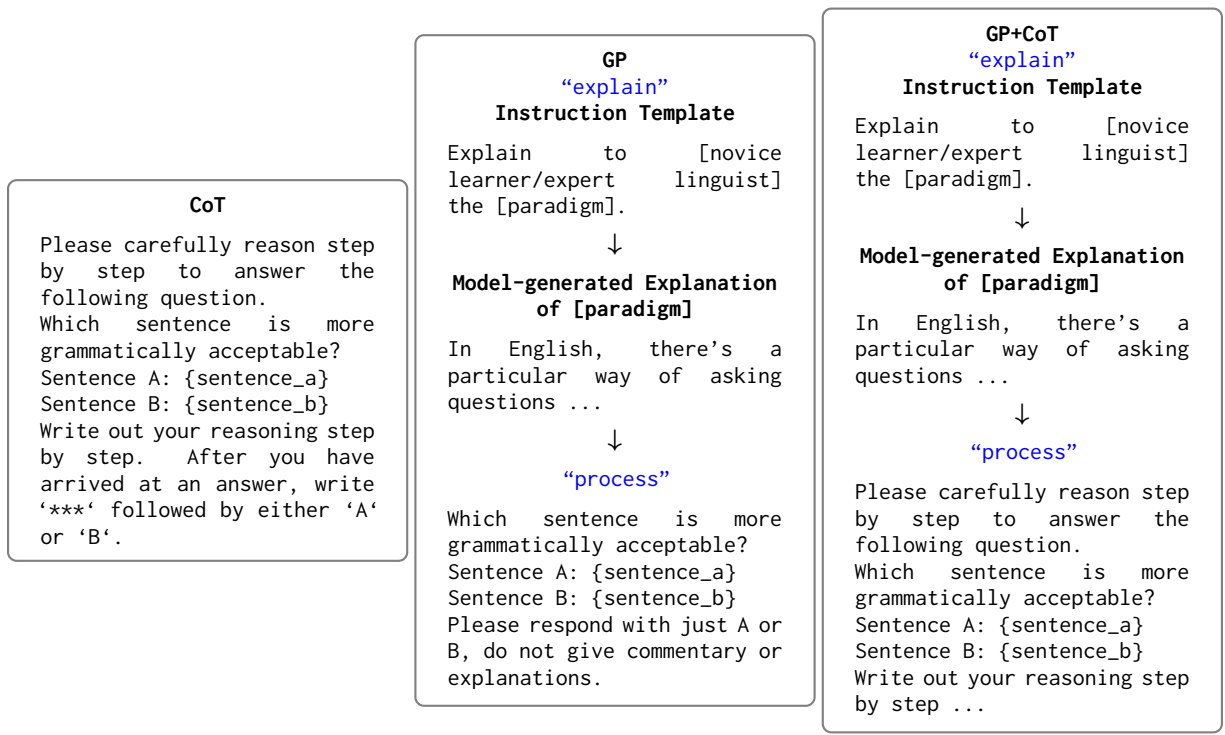

    \centering
    \small
    \noindent
    \begin{minipage}{0.33\textwidth}
    \begin{tcolorbox}[colframe=gray, colback=white, boxrule=0.8pt, fontupper=\ttfamily,boxsep=0.3pt,width=\linewidth]
\begin{center}\textbf{CoT}\end{center}
    Please carefully reason step by step to answer the following question.\\
    Which sentence is more grammatically acceptable? \\
    Sentence A: \texttt{\{sentence\_a\}} \\
    Sentence B: \texttt{\{sentence\_b\}} \\
    Write out your reasoning step by step. After you have arrived at an answer, write `***` followed by either `A` or `B`.
    \end{tcolorbox}
    \end{minipage}%
    \hfill
    \begin{minipage}{0.33\textwidth}
    \begin{tcolorbox}[colframe=gray, colback=white, boxrule=0.8pt, fontupper=\ttfamily,boxsep=0.3pt,width=\linewidth]
    \begin{center}\textbf{GP}\\
\textcolor{blue}{``explain''}\\
    \textbf{Instruction Template}\end{center}
    Explain to [novice learner/expert linguist] the [paradigm].
\begin{center}$\boldsymbol\downarrow$\end{center}
\begin{center}\textbf{Model-generated Explanation of [paradigm]}\end{center}
In English, there's a particular way of asking questions ...
\begin{center}$\boldsymbol\downarrow$\end{center}
\begin{center}\textcolor{blue}{``process''}\end{center}
    Which sentence is more grammatically acceptable? \\
    Sentence A: \texttt{\{sentence\_a\}} \\
    Sentence B: \texttt{\{sentence\_b\}} \\
    Please respond with just A or B, do not give commentary
or explanations.
    \end{tcolorbox}
    \end{minipage}%
    \hfill
    \begin{minipage}{0.33\textwidth}
    \begin{tcolorbox}[colframe=gray, colback=white, boxrule=0.8pt, fontupper=\ttfamily,boxsep=0.3pt,width=\linewidth]
\begin{center}\textbf{GP+CoT}\\
\textcolor{blue}{``explain''}\\
    \textbf{Instruction Template}\end{center}
    Explain to [novice learner/expert linguist] the [paradigm].
\begin{center}$\boldsymbol\downarrow$\end{center}
\begin{center}\textbf{Model-generated Explanation of [paradigm]}\end{center}
In English, there's a particular way of asking questions ...
\begin{center}$\boldsymbol\downarrow$\end{center}
\begin{center}\textcolor{blue}{``process''}\end{center}
Please carefully reason step by step to answer the following question.\\
    Which sentence is more grammatically acceptable? \\
    Sentence A: \texttt{\{sentence\_a\}} \\
    Sentence B: \texttt{\{sentence\_b\}} \\
    Write out your reasoning step by step ...
    \end{tcolorbox}
    \end{minipage}
    \caption{An illustration of Chain-of-Thought (CoT) reasoning vs. Grammar Prompting (GP) vs. Grammar Prompting with Chain-of-Thought reasoning (GP+CoT)}
    \label{fig:overview}
\end{figure*}

We observed this pattern across different LLMs and linguistic phenomena: when making acceptability judgments, models typically begin by paraphrasing (or translating) the sentences, and in this process frequently obscure the grammatical features crucial for accurate judgments. 

How can we help models focus more directly on linguistic structure? To explore this question, we adopt a classic psycholinguistics task: \textit{grammatical acceptability judgments} \cite{schuetze_t2016empirical}.
Specifically, we evaluate whether models can reliably distinguish grammatically acceptable forms from grammatically unacceptable forms by using minimal pairs that differ by only one syntactic feature (e.g., a missing question particle). 
We focus on  three typologically distinct languages: English, Chinese, and Russian. 

Following \citet{mahowald2024dissociatinglanguagethoughtlarge}, we distinguish between two kinds of linguistic competence: \textit{formal} -- the knowledge of structural rules needed to differentiate grammatical from ungrammatical forms -- and \textit{functional} -- practical language use in tasks like translation or summarization. While language models often display strong functional competence, they may possess implicit syntactic knowledge but fail to access it systematically, particularly in grammatical acceptability judgments.

We introduce the \textbf{explain-then-process} 
paradigm for grammar prompting (see Figure~\ref{fig:overview}), which has two steps: first, a grammatical explanation of a given paradigm is elicited from an LLM; then the grammatical explanation is supplied { back to an LLM or a smaller language model (SLM)},
and we investigate the interplay between grammar prompting and chain-of-thought reasoning. By equipping LLMs {and SLMs} with these targeted prompts, we find that models more consistently identify syntactic violations, suggesting that explicit grammar guidance can unlock more systematic linguistic reasoning. 


We make the following contributions: 

\vspace{-2mm}
\begin{itemize} 
\item A novel explain-then-process paradigm that elicits explanations from models for a given task before using them for processing.
\vspace{-2mm}
\item A grammar prompting method that leverages explicit metalinguistic input to further unlock a model's linguistic reasoning and boost accuracy in minimal-pair grammatical acceptability task, {especially for SLMs}.
\vspace{-2mm}
\item A systematic evaluation of multiple language models from the GPT, Claude, Llama families across English, Chinese, and Russian, analyzing how well grammar prompting improves their ability to make structured linguistic judgments.

\end{itemize}

\section{Related Work}
\label{sec:related_work}

Recent theoretical work by \citet{wies2024learnability} suggests that LLMs possess considerable latent linguistic knowledge that can be activated through in-context learning (ICL) and introducing new linguistic information. For example, \citet{lampinen-etal-2022-language} and \citet{srivastava2023imitationgamequantifyingextrapolating} demonstrate the effectiveness of targeted explanations and validation-tuned prompts to improve ability in specific tasks. Notably, the \textit{Machine Translation from One Book} (MTOB) paradigm  found that LLMs could improve translation between English and a zero-resource Papuan language, Kalamang, by consulting one grammar book \cite{tanzer2024a}. Follow-up work showed that incorporating explicit rules can help in some cases \cite{guo-etal-2024-teaching, zhang-etal-2024-teaching}, but might be insufficient without additional strategies such as validation-tuned prompts \cite{lampinen-etal-2022-language, srivastava2023imitationgamequantifyingextrapolating}.

{Grammatical acceptability judgment tasks have long been a fundamental method in psycholinguistics for assessing linguistic competence, typically presenting single sentences for evaluation \cite{schuetze_t2016empirical}. In computational linguistics, this task has been systematized using minimal pairs--sentence pairs that differ by only one syntactic feature--allowing for more controlled comparisons. This approach led to the development of benchmark datasets such as the English-language CoLA and BLiMP \cite{warstadt2019Acceptability, warstadt-etal-2020-blimp-benchmark}, which have inspired similar resources for other languages, including Chinese \cite{song-etal-2022-sling}, Russian \cite{taktasheva2024rublimprussianbenchmarklinguistic}, Arabic \cite{alrajhi-etal-2022-assessing}, Dutch \cite{suijkerbuijkblimp}, and Indonesian \cite{leong2023bhasaholisticsoutheastasian}. While some computational approaches to minimal pair evaluation still rely on intrinsic metrics like perplexity or model-assigned probabilities, there is growing interest in prompt-based methods that explicitly ask models to make grammatical acceptability judgments \cite{qiu2024grammaticalityrepresentationchatgptcompared}, aligning more closely with how LLMs are used in interactive settings.}

\section{Explain-then-Process Grammar Prompting} \label{sec:experimental_setup}

We hypothesize that while LLMs can often {describe} grammatical rules, they do not consistently apply these rules when making grammatical acceptability judgments. To address this, we propose an \textit{explain-then-process} framework for grammar prompting allowing us to test whether LLMs can integrate explicit metalinguistic knowledge into their grammatical reasoning. 

\begin{enumerate}[noitemsep]
    \item {\bf explain} - the model first generates a detailed description of a grammatical phenomenon, outlining key syntactic constraints. 
    \item {\bf process} - this explanation is then fed back to the model to guide syntactic decision-making and improve processing.
\end{enumerate}

By explicitly linking grammatical descriptions to acceptability judgments, this approach tests whether LLMs can internalize syntactic rules. This is particularly important in multilingual settings, where access to explicit grammatical explanations is limited for many languages. Unlike English, which benefits from extensive linguistic resources, many languages lack formalized grammatical descriptions. Explain-then-process prompting allows LLMs to self-generate grammatical explanations, reducing reliance on external resources.




To elicit principled and consistent grammatical explanations from LLMs, we first design an {\em instruction template} (Figure~\ref{table_metaprompt} shows a sample template) so as to 
ensures systematic coverage across languages and structures. Each template includes the grammatical paradigm name (e.g., NPI licensing), sample minimal pairs, and an instruction to avoid full example sentences, preventing reasoning based on examples \cite{aycock2024llmsreallylearntranslate}. The instruction template also specifies a target audience: we request either a description for {\em beginners} with clear, straightforward explanations, or for {\em experts} with technical descriptions and formal terminology, following MTOB \cite{tanzer2024a}.






\begin{figure}[!t]

\begin{tcolorbox}[
  colframe=black, colback=white, 
  boxrule=0.8pt, rounded corners, 
  width=\columnwidth, 
  left=0pt, right=0pt, boxsep=1pt
]

\textbf{Task:} Explain, to a [novice learner/expert linguist], the `left branch island echo question' in English.

\textbf{Context:} Help a [novice learner/expert linguist] understand and identify correct usage of this concept to pass a test.

\textbf{Requirements:}
\begin{itemize}[noitemsep,topsep=0pt]
\item Clear, non-technical title
\item Explain key rules and patterns
\item Include specific grammatical details
\item ...
\end{itemize}

\textbf{Examples} (for reference, do not include):\\
Good: Many boys will embarrass what patient?\\
Bad: What will many boys embarrass patient?\\
Good: Sara was insulting what student?\\
Bad: What was Sara insulting student?

\textbf{Target length:} 250 words
\end{tcolorbox}
\caption{Instruction Template for Eliciting Grammar Explanations}
\label{table_metaprompt}\vspace{-0.4cm}
\end{figure}

\begin{figure}[!t]
\begin{tcolorbox}[
  colframe=black, colback=gray!10, 
  boxrule=0.8pt, rounded corners, 
  width=\columnwidth, 
  fontupper=\ttfamily
]
\small 

\textbf{Understanding Special Question Patterns in English}

In English, there's a particular way of asking questions that might seem tricky at first. This pattern involves asking about a specific part of a sentence while keeping most of the original sentence structure intact.

\textbf{Key Rules:}
\begin{enumerate}[noitemsep,topsep=0pt]
\item ... [truncated for brevity]
\end{enumerate}

\textbf{Grammatical Details:}
\begin{itemize}[noitemsep,topsep=0pt]
\item ... [truncated for brevity]
\end{itemize}

\textbf{Tips for Correct Usage:}
\begin{itemize}[noitemsep,topsep=0pt]
\item ... [truncated for brevity]
\end{itemize}

\textbf{To Spot Incorrect Usage:}
\begin{itemize}[noitemsep,topsep=0pt]
\item ... [truncated for brevity]
\end{itemize}

Remember, this question form is used in specific contexts, often to express surprise or seek confirmation. It's not the standard way to form all questions in English, but it's important to recognize and use correctly when appropriate.
\end{tcolorbox}
\vspace{-0.4cm}
\caption{Sonnet-generated `Beginner-Oriented' Grammar Explanation for `\textit{left branch island echo question}'}\label{fig:sample-explanation}
\end{figure}

Next, we use these instructions to ask LLMs to generate targeted grammatical explanations, or {\em grammar prompts}. Figure~\ref{fig:sample-explanation} shows a grammatical explanation generated by Sonnet. Given that the models we examine have English as their dominant training language \cite{wendler-etal-2024-llamas}, we used English  prompts throughout
, while the minimal pair sentences were in the target language.

\section{Experimental Setup}
\subsection{Models}

We evaluated five models, including  top-tier proprietary models such as GPT-4o (\texttt{gpt-4o-2024-08-06}) and Claude 3.5 Sonnet (\texttt{claude-3-5-sonnet-20240620}) and their smaller counterparts such as GPT-3.5 (\texttt{gpt-3.5-turbo-0125}) and Claude 3.5 Haiku (\texttt{claude-3-haiku-20240307}), along with Llama 3.3 9B as an open-source multilingual model.

\subsection{Datasets}
\label{sec:datasets}
We focus our evaluation on three languages. Each minimal pair dataset covers a specific syntactic phenomenon with minimally different sentence pairs -- one grammatical and one ungrammatical. Each dataset organizes its minimal pair paradigms into broad syntactic categories (e.g., island effects, classifier agreement). 


\textbf{\em English} The Benchmark of Linguistic Minimal Pairs (BLiMP) \cite{warstadt-etal-2020-blimp-benchmark} spans 67 syntactic paradigms (e.g., subject-verb agreement, island constraints, negative polarity licensing) that were extracted from English syntax textbooks and cover diverse aspects of English grammar. 
 
\textbf{\em Chinese} The Benchmark of Sino LINGuistics (SLING) \cite{song-etal-2022-sling} comprises treebank-derived syntactic minimal pairs, covering 38 paradigms including phenomena like aspect marking with \textit{le}, \textit{guo} and \textit{zai}, and wh-fronting.


\textbf{\em Russian} The Russian Benchmark of Linguistic Minimal Pairs (RuBLiMP) \cite{taktasheva2024rublimprussianbenchmarklinguistic} is made up of corpus-derived minimal pairs characterized by lexical and syntactic diversity. The dataset comprises 45 paradigms covering morphosyntactic phenomena like subject-predicate agreement and reflexive binding.

\paragraph{Focusing on ``challenging'' paradigms} \quad  Initial testing showed gpt-4o achieving near-perfect accuracy on many paradigms. Appendix \ref{sec:appendix_full_baselines} contains the  full set of results. We selected the most challenging paradigms across 8 categories from BLiMP (90\% or lower accuracy), 6 from SLING (excluding easier relative clause and anaphora categories), and 7 categories from RuBLiMP ($\leq$96\% accuracy). Each category includes 1-5 minimal pair paradigms, and we use the first 50 minimal pairs per paradigm. 


\subsection{Presentation Schema}
\label{sec:presentation-schema}
Intrinsic metrics such as probability or perplexity, often used in minimal pair datasets to probe LLMs' syntactic knowledge \cite{warstadt2019Acceptability, warstadt-etal-2020-blimp-benchmark, alrajhi-etal-2022-assessing, song-etal-2022-sling}, frequently fail to capture how models behave when explicitly asked to judge grammaticality \cite{suijkerbuijkblimp}.
Furthermore, access to intrinsic metrics is limited for proprietary models, making direct prompt-based evaluations of grammatical competence a more practical and usage-aligned alternative as LLMs are deployed in applications requiring explicit linguistic processing and feedback. 

Following \citet{leong2023bhasaholisticsoutheastasian}, we present both sentences of a minimal pair simultaneously. This approach serves to focus the model's attention on the critical grammatical contrast and to avoid yes-response bias, where models show a clear preference toward positive answers \cite{yes-bias-dentella}, which we observed in preliminary tests when presenting single sentences.
Additionally, we systematically vary the presentation order of grammatical and ungrammatical sentences across three A/B prompting trials:
\textbf{Trial 1}: The grammatical sentence is presented as ``Sentence A'' and the ungrammatical as ``Sentence B''.
\textbf{Trial 2}: The order is reversed.
\textbf{Trial 3}: The assignment is randomized.
The results of these trials are averaged.

\subsection{Baseline Conditions}

We include several different types of conditions in our experiments. 

\begin{itemize}
    \item \texttt{Base} is the basic prompt as follows:\\
    \begin{small} {Which sentence is more grammatically acceptable? \\
    \small Sentence A: \texttt{\{sentence\_a\}} \\
    \small Sentence B: \texttt{\{sentence\_b\}} \\
    \small Please respond with just A or B, do not give commentary or explanations.}\end{small}
    
    \item \texttt{Chain-of-Thought} (CoT) asks the model to reason step by step before answering (Fig.~\ref{fig:overview})
    \item \texttt{Control} (con) provides an explanation of an {\em irrelevant} linguistic phenomenon (the ``null quotative'') which does not correspond to any tested paradigms, instead of providing a grammar explanation relevant to the minimal pair.  This control ensures that performance gains are due to targeted grammar explanations rather than generic instruction-following behavior.
    \item \texttt{Textbook} (txt) provides a compilation of multiple grammatical explanations instead of a single relevant grammar prompt, and models must select and apply the correct rule for each minimal pair.
    \item \texttt{3-shot} {encourages superficial pattern matching instead of reasoning \cite{du2023shortcut}, so we do not consider it a strictly comparable baseline; nevertheless, we present and discuss three-shot results in Section \ref{sec:three-shot}}.
\end{itemize}

The proposed grammar prompts -- \texttt{beginner} (\texttt{GPb}) and \texttt{expert} (\texttt{GPx}) -- are tested with both base and CoT prompts. To generate the explanations, we initially leveraged the two strongest available models: Claude 3.5 Sonnet and GPT-4o, but we transitioned from gpt-4o to GPT-o1 when it was released. Small-scale manual verification confirmed the quality and adherence to the instructions.



\begin{table}[!t]
\centering
\small
\begin{tabular}{@{}l S[table-format=2.1]S[table-format=2.1] | S[table-format=2.1]S[table-format=2.1]@{}}
\toprule
& \multicolumn{2}{c}{gpt-3.5} & \multicolumn{2}{c}{gpt-4o} \\
Categories & \multicolumn{1}{c}{{GPb}} & \multicolumn{1}{c}{GPx} & \multicolumn{1}{c}{GPb} & \multicolumn{1}{c}{GPx} \\
\midrule
\multicolumn{5}{c}{\cellcolor{gray!20}\textsc{English \textbf{(BLIMP)}}} \\

NPI lic. & \multicolumn{1}{c}{{69.7}} & \multicolumn{1}{c}{\textbf{75.3}} & \multicolumn{1}{c}{{87.7}} & \multicolumn{1}{c}{\textbf{91.3}} \\
arg. Str. & \multicolumn{1}{c}{\textbf{79.3}} & \multicolumn{1}{c}{{68.0}} & \multicolumn{1}{c}{\textbf{90.7}} & \multicolumn{1}{c}{{88.0}} \\
binding & \multicolumn{1}{c}{\textbf{67.7}} & \multicolumn{1}{c}{{63.3}} & \multicolumn{1}{c}{\textbf{80.3}} & \multicolumn{1}{c}{{67.0}} \\
cntrl/Rs. & \multicolumn{1}{c}{\textbf{92.9}} & \multicolumn{1}{c}{{85.3}} & \multicolumn{1}{c}{\textbf{92.4}} & \multicolumn{1}{c}{{91.1}} \\
ellipsis & \multicolumn{1}{c}{{72.7}} & \multicolumn{1}{c}{\textbf{76.0}} & \multicolumn{1}{c}{\textbf{69.7}} & \multicolumn{1}{c}{{58.3}} \\
filler gap & \multicolumn{1}{c}{{58.9}} & \multicolumn{1}{c}{\textbf{78.2}} & \multicolumn{1}{c}{{81.3}} & \multicolumn{1}{c}{\textbf{91.1}} \\
island ef. & \multicolumn{1}{c}{\textbf{65.1}} & \multicolumn{1}{c}{{57.5}} & \multicolumn{1}{c}{\textbf{79.1}} & \multicolumn{1}{c}{{72.0}} \\
quantifiers & \multicolumn{1}{c}{{83.0}} & \multicolumn{1}{c}{\textbf{86.3}} & \multicolumn{1}{c}{\textbf{100.0}} & \multicolumn{1}{c}{{90.3}} \\
\noalign{\smallskip}\hdashline \noalign{\smallskip}
average (BLiMP) & \multicolumn{1}{c}{73.6} & \multicolumn{1}{c}{\textbf{73.7}} & \multicolumn{1}{c}{\textbf{85.15}} & \multicolumn{1}{c}{{81.14}} \\
\multicolumn{5}{c}{\cellcolor{gray!20}\textsc{{Chinese \textbf{(SLING)}}}} \\

alt. quest. & \multicolumn{1}{c}{\textbf{100.0}} & \multicolumn{1}{c}{\textbf{100.0}} & \multicolumn{1}{c}{\textbf{100.0}} & \multicolumn{1}{c}{\textbf{100.0}} \\
aspect & \multicolumn{1}{c}{\textbf{78.8}} & \multicolumn{1}{c}{{76.5}} & \multicolumn{1}{c}{\textbf{99.0}} & \multicolumn{1}{c}{{97.0}} \\
classifier-Noun & \multicolumn{1}{c}{\textbf{75.1}} & \multicolumn{1}{c}{{72.1}} & \multicolumn{1}{c}{\textbf{89.9}} & \multicolumn{1}{c}{{88.0}} \\
definiteness & \multicolumn{1}{c}{\textbf{98.7}} & \multicolumn{1}{c}{{98.2}} & \multicolumn{1}{c}{{99.8}} & \multicolumn{1}{c}{\textbf{100.0}} \\
\noalign{\smallskip}\hdashline \noalign{\smallskip} 
average (SLING) & \multicolumn{1}{c}{\textbf{88.1}} & \multicolumn{1}{c}{{86.7}} & \multicolumn{1}{c}{\textbf{97.1}} & \multicolumn{1}{c}{{96.2}} \\
\midrule
\multicolumn{5}{c}{\cellcolor{gray!20}\textsc{Russian \textbf{(RUBLIMP)}}} \\

anaphor & \multicolumn{1}{c}{\textbf{88.7}} & \multicolumn{1}{c}{{85.3}} & \multicolumn{1}{c}{{93.3}} & \multicolumn{1}{c}{\textbf{94.0}} \\
reflexives & \multicolumn{1}{c}{\textbf{100.0}} & \multicolumn{1}{c}{\textbf{100.0}} & \multicolumn{1}{c}{\textbf{100.0}} & \multicolumn{1}{c}{\textbf{100.0}} \\
subj.-pred. agr & \multicolumn{1}{c}{\textbf{83.8}} & \multicolumn{1}{c}{{83.3}} & \multicolumn{1}{c}{{96.7}} & \multicolumn{1}{c}{\textbf{96.9}} \\
aspect & \multicolumn{1}{c}{\textbf{85.8}} & \multicolumn{1}{c}{{75.8}} & \multicolumn{1}{c}{\textbf{100.0}} & \multicolumn{1}{c}{\textbf{100.0}} \\
government & \multicolumn{1}{c}{{78.0}} & \multicolumn{1}{c}{\textbf{80.0}} & \multicolumn{1}{c}{\textbf{98.3}} & \multicolumn{1}{c}{{96.3}} \\
word formation & \multicolumn{1}{c}{\textbf{98.7}} & \multicolumn{1}{c}{{93.3}} & \multicolumn{1}{c}{\textbf{100.0}} & \multicolumn{1}{c}{\textbf{100.0}} \\

\noalign{\smallskip}\hdashline \noalign{\smallskip}
average (RuBLiMP) & \multicolumn{1}{c}{\textbf{89.1}} & \multicolumn{1}{c}{{86.2}} & \multicolumn{1}{c}{\textbf{98.0}} & \multicolumn{1}{c}{{97.8}} \\
\midrule

average over all & \multicolumn{1}{c}{\textbf{82.0}} & \multicolumn{1}{c}{{80.8}} & \multicolumn{1}{c}{\textbf{92.1}} & \multicolumn{1}{c}{{90.1}} \\
\bottomrule
\end{tabular}
\caption{Results of Beginner (GPb) and Expert (GPx) grammar prompting conditions; explanations generated using \texttt{Claude Sonnet}}
\label{tab:comparison-exp-beg}
\end{table}

\begin{table}[!t]
\centering
\small
\begin{tabular}{@{}l S[table-format=2.1]S[table-format=2.1]S[table-format=2.1]  S[table-format=2.1]S[table-format=2.1]S[table-format=2.1]@{}}
\toprule
 & \multicolumn{3}{c}{gpt-3.5} & \multicolumn{3}{c}{gpt-4o} \\
Categories & \multicolumn{1}{c}{con} & \multicolumn{1}{c}{txt} & \multicolumn{1}{c}{GPb} & \multicolumn{1}{c}{con} & \multicolumn{1}{c}{txt} & \multicolumn{1}{c}{GPb} \\
\midrule
NPI lic. & \multicolumn{1}{c}{{64.0}} & 59.3 & \multicolumn{1}{c}{\textbf{67.3}} & 74.3 & \multicolumn{1}{c}{{81.0}} & \multicolumn{1}{c}{\textbf{93.6}} \\
arg. str. & \multicolumn{1}{c}{\textbf{69.7}} & \multicolumn{1}{c}{{65.3}} & 46.7 & 84.3 & \multicolumn{1}{c}{\textbf{90.3}} & \multicolumn{1}{c}{{89.6}} \\
binding & \multicolumn{1}{c}{\textbf{57.5}} & 56.0 & \multicolumn{1}{c}{{57.3}} & \multicolumn{1}{c}{{74.8}} & 73.7 & \multicolumn{1}{c}{\textbf{78.0}} \\
cntrl/rs. & 75.4 & \multicolumn{1}{c}{{75.6}} & \multicolumn{1}{c}{\textbf{87.8}} & 81.6 & \multicolumn{1}{c}{{85.8}} & \multicolumn{1}{c}{\textbf{94.7}} \\
ellipsis & \multicolumn{1}{c}{{69.5}} & 58.2 & \multicolumn{1}{c}{\textbf{74.0}} & \multicolumn{1}{c}{{69.7}} & 69.5 & \multicolumn{1}{c}{\textbf{89.5}} \\
filler gap & 46.9 & \multicolumn{1}{c}{{53.4}} & \multicolumn{1}{c}{\textbf{88.9}} & 70.0 & \multicolumn{1}{c}{{73.8}} & \multicolumn{1}{c}{\textbf{95.1}} \\
island ef. & 63.0 & \multicolumn{1}{c}{\textbf{63.3}} & \multicolumn{1}{c}{{63.1}} & 71.6 & \multicolumn{1}{c}{{75.4}} & \multicolumn{1}{c}{\textbf{81.6}} \\
quantifiers & \multicolumn{1}{c}{{66.7}} & 59.0 & \multicolumn{1}{c}{\textbf{95.0}} & \multicolumn{1}{c}{{80.2}} & 73.0 & \multicolumn{1}{c}{\textbf{99.2}} \\
\midrule
average & \multicolumn{1}{c}{{64.1}} & 61.3 & \multicolumn{1}{c}{\textbf{72.5}} & 75.8 & \multicolumn{1}{c}{{77.8}} & \multicolumn{1}{c}{\textbf{90.2}} \\
\bottomrule
\end{tabular}
\caption{{Control} and {Textbook} condition comparison on BLiMP dataset}
\label{tab:control_setting_summary}
\end{table}

\section{Results and Analysis}
\label{sec:results}
We present the results (macro-average accuracy across paradigm categories) of four main experiments: (1) comparing beginner vs. expert-oriented grammar prompts (Section \ref{sec:comparison-exp-beg}), (2) evaluating control and textbook conditions (Section \ref{sec:control-textbook-section}), (3) assessing performance on BLiMP, SLING, and RuBLiMP (Sections \ref{sec:blimp-analysis}--\ref{sec:rublimp_analysis}), and (4) comparison with three-shot results (Section \ref{sec:three-shot}). In Section \ref{sec:slm-llm-gap} we examine how grammar prompting narrows the gap between LLM and SLM performance. 

\subsection{Beginner {\em vs.} Expert  Explanations}
\label{sec:comparison-exp-beg}
Table~\ref{tab:comparison-exp-beg} compares `beginner' and `expert' prompts across all datasets using gpt-3.5 and gpt-4o models. In a macro-analysis across 18 categories, simpler beginner-oriented explanations generally outperformed expert-level explanations by a small but significant margin (-1.9\% $\pm$  5.7\%, p=0.002, d=-0.344). Based on these results, we limited further investigations to `beginner'-oriented prompts. 

The prompting styles differ markedly: beginner prompts emphasize practical recognition (``Use `who' or `what' when the following part of the sentence is missing information''), while expert prompts employ technical terminology (``long-distance dependencies,'' ``selectional restrictions''). Based on these results, we limited further investigations to beginner-oriented prompts. Additional examples are included in Appendix \ref{sec:instructiontemplate}.


\begin{table*}[!t]
\centering
\small
\setlength{\tabcolsep}{4pt}
\begin{tabular}{lcccccc|cccccc}
\toprule
& \multicolumn{6}{c}{gpt-3.5} & \multicolumn{6}{c}{Haiku} \\
Categories & \multicolumn{1}{c}{\text{base}} & \multicolumn{1}{c}{\text{CT}} & \multicolumn{1}{c}{\text{GPb}} & \multicolumn{1}{c}{\text{GPb}} & \multicolumn{1}{c}{\text{GPb+CT}} & \multicolumn{1}{c}{\text{GPb+CT}} & \multicolumn{1}{c}{\text{base}} & \multicolumn{1}{c}{\text{CT}} & \multicolumn{1}{c}{\text{GPb}} & \multicolumn{1}{c}{\text{GPb}} & \multicolumn{1}{c}{\text{GPb+CT}} & \multicolumn{1}{c}{\text{GPb+CT}} \\
 & \multicolumn{1}{c}{\text{}} & \multicolumn{1}{c}{\text{}} & \multicolumn{1}{c}{\text{son}} & \multicolumn{1}{c}{\text{o1}} & \multicolumn{1}{c}{\text{son}} & \multicolumn{1}{c}{\text{o1}} & \multicolumn{1}{c}{\text{}} & \multicolumn{1}{c}{\text{}} & \multicolumn{1}{c}{\text{son}} & \multicolumn{1}{c}{\text{o1}} & \multicolumn{1}{c}{\text{son}} & \multicolumn{1}{c}{\text{o1}} \\
\midrule
NPI licensing (2) & 60.0 & 67.3 & 69.7 & 67.3 & \multicolumn{1}{c}{\textbf{88.7}} & \multicolumn{1}{c}{\underline{87.0}} & 57.3 & 73.3 & 77.0 & 65.3 & \multicolumn{1}{c}{\underline{96.0}} & \multicolumn{1}{c}{\textbf{96.7}} \\
arg. struct. (1) & \multicolumn{1}{c}{\underline{77.3}} & 66.0 & \multicolumn{1}{c}{\textbf{79.3}} & 46.7 & 74.7 & 66.0 & 74.7 & 85.3 & \multicolumn{1}{c}{\textbf{86.7}} & 74.0 & 82.7 & \multicolumn{1}{c}{\underline{86.0}} \\
binding (2) & 65.3 & 57.3 & \multicolumn{1}{c}{\underline{67.7}} & 57.3 & \multicolumn{1}{c}{\textbf{68.7}} & 62.0 & 56.7 & 63.7 & 64.7 & 50.7 & \multicolumn{1}{c}{\textbf{74.3}} & \multicolumn{1}{c}{\underline{70.7}} \\
control/raising (3) & 83.1 & 79.1 & \multicolumn{1}{c}{\underline{92.9}} & 87.8 & \multicolumn{1}{c}{\textbf{94.4}} & 82.9 & 69.3 & 83.1 & 75.1 & 70.7 & \multicolumn{1}{c}{\textbf{94.2}} & \multicolumn{1}{c}{\underline{90.4}} \\
ellipsis (2) & \multicolumn{1}{c}{\underline{76.0}} & 69.7 & 72.7 & 74.0 & 72.0 & \multicolumn{1}{c}{\textbf{79.7}} & 60.3 & \multicolumn{1}{c}{\underline{75.0}} & 59.3 & 50.0 & 73.7 & \multicolumn{1}{c}{\textbf{87.3}} \\
filler gap (3) & 52.7 & 48.4 & 58.9 & \multicolumn{1}{c}{\textbf{88.9}} & 59.6 & \multicolumn{1}{c}{\underline{86.0}} & 55.6 & 62.9 & 62.4 & \multicolumn{1}{c}{\underline{87.6}} & 73.1 & \multicolumn{1}{c}{\textbf{94.7}} \\
island effect (5) & 68.9 & 61.6 & 65.1 & 63.1 & \multicolumn{1}{c}{\textbf{74.4}} & \multicolumn{1}{c}{\underline{70.9}} & 60.9 & 64.7 & 60.5 & 59.1 & \multicolumn{1}{c}{\underline{73.3}} & \multicolumn{1}{c}{\textbf{74.3}} \\
quantifiers (2) & 59.7 & 52.0 & 83.0 & \multicolumn{1}{c}{\textbf{95.0}} & 91.0 & \multicolumn{1}{c}{\underline{92.7}} & 54.7 & 68.3 & 79.0 & 81.3 & \multicolumn{1}{c}{\underline{91.3}} & \multicolumn{1}{c}{\textbf{91.7}} \\
\midrule
Average & 67.9 & 62.7 & 73.6 & 72.5 & \multicolumn{1}{c}{\underline{77.9}} & \multicolumn{1}{c}{\textbf{78.4}} & 61.2 & 72.0 & 70.6 & 67.3 & \multicolumn{1}{c}{\underline{82.3}} & \multicolumn{1}{c}{\textbf{86.5}} \\
\midrule
& \multicolumn{6}{c}{gpt-4o} & \multicolumn{6}{c}{Sonnet} \\
Categories & \multicolumn{1}{c}{\text{base}} & \multicolumn{1}{c}{\text{CT}} & \multicolumn{1}{c}{\text{GPb}} & \multicolumn{1}{c}{\text{GPb}} & \multicolumn{1}{c}{\text{GPb+CT}} & \multicolumn{1}{c}{\text{GPb+CT}} & \multicolumn{1}{c}{\text{base}} & \multicolumn{1}{c}{\text{CT}} & \multicolumn{1}{c}{\text{GPb}} & \multicolumn{1}{c}{\text{GPb}} & \multicolumn{1}{c}{\text{GPb+CT}} & \multicolumn{1}{c}{\text{GPb+CT}} \\
 & \multicolumn{1}{c}{\text{}} & \multicolumn{1}{c}{\text{}} & \multicolumn{1}{c}{\text{son}} & \multicolumn{1}{c}{\text{o1}} & \multicolumn{1}{c}{\text{son}} & \multicolumn{1}{c}{\text{o1}} & \multicolumn{1}{c}{\text{}} & \multicolumn{1}{c}{\text{}} & \multicolumn{1}{c}{\text{son}} & \multicolumn{1}{c}{\text{o1}} & \multicolumn{1}{c}{\text{son}} & \multicolumn{1}{c}{\text{o1}} \\
\midrule
NPI licensing (2) & 73.3 & 81.7 & 87.7 & 93.7 & \multicolumn{1}{c}{\textbf{98.7}} & \multicolumn{1}{c}{\underline{97.0}} & 75.7 & 79.0 & 96.7 & \multicolumn{1}{c}{\underline{99.3}} & \multicolumn{1}{c}{\textbf{100.0}} & \multicolumn{1}{c}{\underline{99.3}} \\
arg. struct. (1) & 88.0 & 88.0 & \multicolumn{1}{c}{\textbf{90.7}} & 86.7 & 84.7 & \multicolumn{1}{c}{\underline{88.7}} & 95.3 & 95.3 & \multicolumn{1}{c}{\underline{96.0}} & 88.7 & \multicolumn{1}{c}{\underline{96.0}} & \multicolumn{1}{c}{\textbf{96.7}} \\
binding (2) & 72.7 & 90.0 & 80.3 & 77.7 & \multicolumn{1}{c}{\textbf{97.7}} & \multicolumn{1}{c}{\underline{92.3}} & 90.7 & 91.0 & \multicolumn{1}{c}{\underline{98.0}} & 91.0 & \multicolumn{1}{c}{\underline{98.0}} & \multicolumn{1}{c}{\textbf{99.0}} \\
control/raising (3) & 86.0 & 89.8 & 92.4 & \multicolumn{1}{c}{\underline{94.9}} & \multicolumn{1}{c}{\textbf{97.3}} & 94.7 & 84.7 & 88.2 & 93.8 & 93.8 & \multicolumn{1}{c}{\textbf{98.4}} & \multicolumn{1}{c}{\underline{95.8}} \\
ellipsis (2) & 70.3 & 85.7 & 69.7 & \multicolumn{1}{c}{\underline{88.7}} & 59.3 & \multicolumn{1}{c}{\textbf{92.0}} & 81.0 & \multicolumn{1}{c}{\underline{88.0}} & 77.0 & \multicolumn{1}{c}{\textbf{98.0}} & 62.0 & \multicolumn{1}{c}{\textbf{98.0}} \\
filler gap (3) & 75.3 & 80.7 & 81.3 & \multicolumn{1}{c}{\underline{94.4}} & 84.7 & \multicolumn{1}{c}{\textbf{95.6}} & 80.2 & 71.3 & 90.4 & \multicolumn{1}{c}{\underline{97.6}} & 85.6 & \multicolumn{1}{c}{\textbf{97.8}} \\
island effect (5) & 77.2 & 77.1 & 79.1 & \multicolumn{1}{c}{\textbf{82.1}} & 80.5 & \multicolumn{1}{c}{\underline{81.6}} & 68.8 & 65.9 & 72.8 & 71.6 & \multicolumn{1}{c}{\textbf{81.3}} & \multicolumn{1}{c}{\underline{74.9}} \\
quantifiers (2) & 81.0 & 86.3 & \multicolumn{1}{c}{\textbf{100.0}} & \multicolumn{1}{c}{\underline{99.3}} & \multicolumn{1}{c}{\textbf{100.0}} & \multicolumn{1}{c}{\textbf{100.0}} & 62.3 & 63.7 & 97.7 & \multicolumn{1}{c}{\underline{98.0}} & \multicolumn{1}{c}{\textbf{100.0}} & \multicolumn{1}{c}{\textbf{100.0}} \\
\midrule
Average & 78.0 & 84.9 & 85.1 & \multicolumn{1}{c}{\underline{89.7}} & 87.9 & \multicolumn{1}{c}{\textbf{92.7}} & 79.8 & 80.3 & 90.3 & \multicolumn{1}{c}{\underline{92.2}} & 90.2 & \multicolumn{1}{c}{\textbf{95.2}} \\
\bottomrule
\end{tabular}
\caption{Different prompting conditions for English-language BLiMP dataset. GPb = beginner-oriented grammar prompting, CT = Chain of Thought. For GP conditions, `son' or `o1'  indicates grammar prompts generated by Claude 3.5 Sonnet or GPT-o1, respectively.}
\label{tab:BLIMP_main_table}
\end{table*}

\subsection{Control {\em vs.} Textbook Conditions}
\label{sec:control-textbook-section}

Table \ref{tab:control_setting_summary} shows the results of two conditions we devised to understand how models utilize grammar prompts. In the \textbf{control} condition, we tested gpt-3.5 and gpt-4o models on BLiMP paradigms using a mismatched grammar prompt (explaining the ``null quotative'' - a syntactic phenomenon irrelevant to any tested minimal pairs). Performance was consistently worse than with GP-beginner's relevant explanations, indicating that models rely on the content of explanations rather than merely responding to their presence. In the \textbf{textbook} condition, we provided a compilation of \textit{all} BLiMP-paradigm beginner grammar prompts as a simulated `syntax textbook'. Performance was similar to the control condition, suggesting that models struggle to select and apply appropriate explanations when presented with too much information.

\begin{table*}[!t]
\centering
\small
\setlength{\tabcolsep}{4pt}
\begin{tabular}{@{}l S[table-format=2.1]S[table-format=2.1]S[table-format=2.1]S[table-format=2.1]S[table-format=2.1]S[table-format=2.1] | S[table-format=2.1]S[table-format=2.1]S[table-format=2.1]S[table-format=2.1]S[table-format=2.1]S[table-format=2.1]@{}}
\toprule
& \multicolumn{6}{c}{gpt-3.5} & \multicolumn{6}{c}{Haiku} \\
Categories & \multicolumn{1}{c}{\text{base}} & \multicolumn{1}{c}{\text{CT}} & \multicolumn{1}{c}{\text{GPb}} & \multicolumn{1}{c}{\text{GPb}} & \multicolumn{1}{c}{\text{GPb+CT}} & \multicolumn{1}{c}{\text{GPb+CT}} & \multicolumn{1}{c}{\text{base}} & \multicolumn{1}{c}{\text{CT}} & \multicolumn{1}{c}{\text{GPb}} & \multicolumn{1}{c}{\text{GPb}} & \multicolumn{1}{c}{\text{GPb+CT}} & \multicolumn{1}{c}{\text{GPb+CT}} \\
 & \multicolumn{1}{c}{\text{}} & \multicolumn{1}{c}{\text{}} & \multicolumn{1}{c}{\text{son}} & \multicolumn{1}{c}{\text{o1}} & \multicolumn{1}{c}{\text{son}} & \multicolumn{1}{c}{\text{o1}} & \multicolumn{1}{c}{\text{}} & \multicolumn{1}{c}{\text{}} & \multicolumn{1}{c}{\text{son}} & \multicolumn{1}{c}{\text{o1}} & \multicolumn{1}{c}{\text{son}} & \multicolumn{1}{c}{\text{o1}} \\
\midrule
altern. quest. (1) & 18.7 & 16.7 & \multicolumn{1}{c}{\textbf{100.0}} & \multicolumn{1}{c}{\underline{99.3}} & \multicolumn{1}{c}{\textbf{100.0}} & 88.7 & \multicolumn{1}{c}{\underline{46.0}} & 31.3 & \multicolumn{1}{c}{\textbf{100.0}} & \multicolumn{1}{c}{\textbf{100.0}} & \multicolumn{1}{c}{\textbf{100.0}} & \multicolumn{1}{c}{\textbf{100.0}} \\
Aspect (5) & 73.2 & 76.9 & 83.1 & 88.7 & \multicolumn{1}{c}{\textbf{94.1}} & \multicolumn{1}{c}{\underline{93.9}} & 77.9 & 89.2 & 95.5 & \multicolumn{1}{c}{\underline{96.1}} & \multicolumn{1}{c}{\textbf{98.9}} & \multicolumn{1}{c}{\textbf{98.9}} \\
Classifier-Noun (6) & \multicolumn{1}{c}{\underline{82.7}} & 81.9 & 75.1 & 72.1 & 81.0 & \multicolumn{1}{c}{\textbf{82.8}} & 84.0 & \multicolumn{1}{c}{\textbf{86.3}} & 68.6 & 63.7 & 81.1 & \multicolumn{1}{c}{\underline{85.4}} \\
Definiteness (3) & \multicolumn{1}{c}{\underline{97.6}} & 89.8 & \multicolumn{1}{c}{\textbf{98.7}} & 94.2 & \multicolumn{1}{c}{\textbf{98.7}} & 86.0 & 88.7 & 95.8 & 96.0 & 97.1 & \multicolumn{1}{c}{\underline{98.0}} & \multicolumn{1}{c}{\textbf{99.1}} \\
Polarity Item (3) & 73.8 & 83.1 & 96.7 & 98.7 & \multicolumn{1}{c}{\underline{99.1}} & \multicolumn{1}{c}{\textbf{99.8}} & 75.1 & 79.8 & \multicolumn{1}{c}{\textbf{100.0}} & \multicolumn{1}{c}{\underline{99.6}} & \multicolumn{1}{c}{\textbf{100.0}} & \multicolumn{1}{c}{\textbf{100.0}} \\
Wh Fronting (2) & \multicolumn{1}{c}{\textbf{98.3}} & 90.3 & 83.0 & 83.0 & \multicolumn{1}{c}{\underline{91.7}} & 88.7 & 84.0 & 99.0 & \multicolumn{1}{c}{\textbf{100.0}} & \multicolumn{1}{c}{\underline{99.7}} & \multicolumn{1}{c}{\textbf{100.0}} & \multicolumn{1}{c}{\textbf{100.0}} \\
\midrule
Average & 74.0 & 73.1 & 89.4 & 89.3 & \multicolumn{1}{c}{\textbf{94.1}} & \multicolumn{1}{c}{\underline{90.0}} & 75.9 & 80.2 & 93.3 & 92.7 & \multicolumn{1}{c}{\underline{96.3}} & \multicolumn{1}{c}{\textbf{97.2}} \\
\midrule
& \multicolumn{6}{c}{gpt-4o} & \multicolumn{6}{c}{Sonnet} \\
Categories & \multicolumn{1}{c}{\text{base}} & \multicolumn{1}{c}{\text{CT}} & \multicolumn{1}{c}{\text{GPb}} & \multicolumn{1}{c}{\text{GPb}} & \multicolumn{1}{c}{\text{GPb+CT}} & \multicolumn{1}{c}{\text{GPb+CT}} & \multicolumn{1}{c}{\text{base}} & \multicolumn{1}{c}{\text{CT}} & \multicolumn{1}{c}{\text{GPb}} & \multicolumn{1}{c}{\text{GPb}} & \multicolumn{1}{c}{\text{GPb+CT}} & \multicolumn{1}{c}{\text{GPb+CT}} \\
 & \multicolumn{1}{c}{\text{}} & \multicolumn{1}{c}{\text{}} & \multicolumn{1}{c}{\text{son}} & \multicolumn{1}{c}{\text{o1}} & \multicolumn{1}{c}{\text{son}} & \multicolumn{1}{c}{\text{o1}} & \multicolumn{1}{c}{\text{}} & \multicolumn{1}{c}{\text{}} & \multicolumn{1}{c}{\text{son}} & \multicolumn{1}{c}{\text{o1}} & \multicolumn{1}{c}{\text{son}} & \multicolumn{1}{c}{\text{o1}} \\
\midrule
altern. quest. (1) & \multicolumn{1}{c}{\underline{92.7}} & 82.7 & \multicolumn{1}{c}{\textbf{100.0}} & \multicolumn{1}{c}{\textbf{100.0}} & \multicolumn{1}{c}{\textbf{100.0}} & \multicolumn{1}{c}{\textbf{100.0}} & \multicolumn{1}{c}{\underline{60.0}} & \multicolumn{1}{c}{\textbf{100.0}} & \multicolumn{1}{c}{\textbf{100.0}} & \multicolumn{1}{c}{\textbf{100.0}} & \multicolumn{1}{c}{\textbf{100.0}} & \multicolumn{1}{c}{\textbf{100.0}} \\
Aspect (5) & 94.1 & 98.1 & 99.2 & \multicolumn{1}{c}{\textbf{99.7}} & \multicolumn{1}{c}{\textbf{99.7}} & \multicolumn{1}{c}{\textbf{99.7}} & 96.5 & 98.8 & 99.5 & 99.3 & \multicolumn{1}{c}{\underline{99.6}} & \multicolumn{1}{c}{\textbf{99.9}} \\
Classifier-Noun (6) & 88.1 & \multicolumn{1}{c}{\textbf{90.0}} & \multicolumn{1}{c}{\underline{89.7}} & 88.4 & 89.4 & 89.6 & 87.4 & \multicolumn{1}{c}{\underline{90.0}} & 84.7 & 86.1 & 89.4 & \multicolumn{1}{c}{\textbf{90.6}} \\
Definiteness (3) & 94.7 & 95.8 & \multicolumn{1}{c}{\underline{99.8}} & \multicolumn{1}{c}{\textbf{100.0}} & 99.6 & \multicolumn{1}{c}{\textbf{100.0}} & 95.6 & 96.9 & \multicolumn{1}{c}{\textbf{100.0}} & \multicolumn{1}{c}{\textbf{100.0}} & \multicolumn{1}{c}{\underline{99.3}} & \multicolumn{1}{c}{\textbf{100.0}} \\
Polarity Item (3) & 89.6 & 95.1 & \multicolumn{1}{c}{\textbf{100.0}} & \multicolumn{1}{c}{\underline{99.6}} & \multicolumn{1}{c}{\textbf{100.0}} & \multicolumn{1}{c}{\textbf{100.0}} & 94.2 & \multicolumn{1}{c}{\underline{95.3}} & \multicolumn{1}{c}{\textbf{100.0}} & \multicolumn{1}{c}{\textbf{100.0}} & \multicolumn{1}{c}{\textbf{100.0}} & \multicolumn{1}{c}{\textbf{100.0}} \\
Wh Fronting (2) & 93.3 & \multicolumn{1}{c}{\underline{98.0}} & \multicolumn{1}{c}{\textbf{100.0}} & \multicolumn{1}{c}{\textbf{100.0}} & \multicolumn{1}{c}{\textbf{100.0}} & \multicolumn{1}{c}{\textbf{100.0}} & \multicolumn{1}{c}{\underline{95.7}} & \multicolumn{1}{c}{\textbf{100.0}} & \multicolumn{1}{c}{\textbf{100.0}} & \multicolumn{1}{c}{\textbf{100.0}} & \multicolumn{1}{c}{\textbf{100.0}} & \multicolumn{1}{c}{\textbf{100.0}} \\
\midrule
Average & 92.1 & 93.3 & 98.1 & 98.0 & \multicolumn{1}{c}{\underline{98.1}} & \multicolumn{1}{c}{\textbf{98.2}} & 88.2 & 96.8 & 97.4 & 97.6 & \multicolumn{1}{c}{\underline{98.1}} & \multicolumn{1}{c}{\textbf{98.4}} \\
\bottomrule
\end{tabular}
\caption{Different prompting conditions for Chinese-language SLING dataset}
\label{tab:SLING_main_table}
\end{table*}


\begin{table*}[t]
\centering
\small
\setlength{\tabcolsep}{4pt}
\begin{tabular}{@{}l S[table-format=2.1]S[table-format=2.1]S[table-format=2.1]S[table-format=2.1]S[table-format=2.1]S[table-format=2.1] | S[table-format=2.1]S[table-format=2.1]S[table-format=2.1]S[table-format=2.1]S[table-format=2.1]S[table-format=2.1]@{}}
\toprule
& \multicolumn{6}{c}{gpt-3.5} & \multicolumn{6}{c}{Haiku} \\
Categories & \multicolumn{1}{c}{\text{base}} & \multicolumn{1}{c}{\text{CT}} & \multicolumn{1}{c}{\text{GPb}} & \multicolumn{1}{c}{\text{GPb}} & \multicolumn{1}{c}{\text{GPb+CT}} & \multicolumn{1}{c}{\text{GPb+CT}} & \multicolumn{1}{c}{\text{base}} & \multicolumn{1}{c}{\text{CT}} & \multicolumn{1}{c}{\text{GPb}} & \multicolumn{1}{c}{\text{GPb}} & \multicolumn{1}{c}{\text{GPb+CT}} & \multicolumn{1}{c}{\text{GPb+CT}} \\
 & \multicolumn{1}{c}{\text{}} & \multicolumn{1}{c}{\text{}} & \multicolumn{1}{c}{\text{son}} & \multicolumn{1}{c}{\text{o1}} & \multicolumn{1}{c}{\text{son}} & \multicolumn{1}{c}{\text{o1}} & \multicolumn{1}{c}{\text{}} & \multicolumn{1}{c}{\text{}} & \multicolumn{1}{c}{\text{son}} & \multicolumn{1}{c}{\text{o1}} & \multicolumn{1}{c}{\text{son}} & \multicolumn{1}{c}{\text{o1}} \\
\midrule
Anaphora (1) & 92.0 & 88.7 & 88.7 & 89.3 & \multicolumn{1}{c}{\underline{92.7}} & \multicolumn{1}{c}{\textbf{93.3}} & 91.3 & 90.7 & 92.7 & 92.7 & \multicolumn{1}{c}{\textbf{96.7}} & \multicolumn{1}{c}{\underline{95.3}} \\
Arg. Str. (1) & 68.7 & \multicolumn{1}{c}{\textbf{77.3}} & 62.0 & \multicolumn{1}{c}{\textbf{77.3}} & 56.7 & \multicolumn{1}{c}{\underline{70.0}} & \multicolumn{1}{c}{\textbf{84.7}} & \multicolumn{1}{c}{\textbf{84.7}} & 73.3 & 79.3 & 74.0 & \multicolumn{1}{c}{\underline{80.7}} \\
Reflexives (1) & 84.7 & 68.0 & \multicolumn{1}{c}{\textbf{100.0}} & \multicolumn{1}{c}{\textbf{100.0}} & \multicolumn{1}{c}{\underline{98.7}} & \multicolumn{1}{c}{\textbf{100.0}} & 90.0 & 90.7 & \multicolumn{1}{c}{\underline{99.3}} & \multicolumn{1}{c}{\textbf{100.0}} & \multicolumn{1}{c}{\textbf{100.0}} & \multicolumn{1}{c}{\underline{99.3}} \\
Subj.-Pred. Agr (5) & 84.1 & 87.3 & 84.8 & 84.5 & \multicolumn{1}{c}{\textbf{92.3}} & \multicolumn{1}{c}{\underline{89.5}} & 90.1 & \multicolumn{1}{c}{\underline{94.0}} & 88.5 & 88.8 & \multicolumn{1}{c}{\underline{94.0}} & \multicolumn{1}{c}{\textbf{94.8}} \\
aspect (3) & 58.9 & 64.7 & 85.8 & 86.4 & \multicolumn{1}{c}{\textbf{97.8}} & \multicolumn{1}{c}{\underline{97.3}} & 63.6 & 73.8 & 97.6 & 88.0 & \multicolumn{1}{c}{\textbf{100.0}} & \multicolumn{1}{c}{\underline{99.3}} \\
government (2) & 74.3 & 69.3 & 78.0 & 69.0 & \multicolumn{1}{c}{\textbf{83.3}} & \multicolumn{1}{c}{\underline{82.7}} & 88.3 & 91.0 & \multicolumn{1}{c}{\underline{96.3}} & 88.7 & \multicolumn{1}{c}{\textbf{97.3}} & 94.0 \\
word formation (1) & 90.0 & 90.0 & \multicolumn{1}{c}{\textbf{98.7}} & \multicolumn{1}{c}{\underline{94.7}} & \multicolumn{1}{c}{\underline{94.7}} & 79.3 & \multicolumn{1}{c}{\underline{98.7}} & \multicolumn{1}{c}{\textbf{100.0}} & \multicolumn{1}{c}{\textbf{100.0}} & 87.3 & \multicolumn{1}{c}{\textbf{100.0}} & 98.0 \\
\midrule
Average & 79.0 & 77.9 & 85.4 & 85.9 & \multicolumn{1}{c}{\textbf{88.0}} & \multicolumn{1}{c}{\underline{87.4}} & 86.7 & 89.3 & 92.5 & 89.3 & \multicolumn{1}{c}{\textbf{94.6}} & \multicolumn{1}{c}{\underline{94.5}} \\
\midrule
& \multicolumn{6}{c}{gpt-4o} & \multicolumn{6}{c}{Sonnet} \\
Categories & \multicolumn{1}{c}{\text{base}} & \multicolumn{1}{c}{\text{CT}} & \multicolumn{1}{c}{\text{GPb}} & \multicolumn{1}{c}{\text{GPb}} & \multicolumn{1}{c}{\text{GPb+CT}} & \multicolumn{1}{c}{\text{GPb+CT}} & \multicolumn{1}{c}{\text{base}} & \multicolumn{1}{c}{\text{CT}} & \multicolumn{1}{c}{\text{GPb}} & \multicolumn{1}{c}{\text{GPb}} & \multicolumn{1}{c}{\text{GPb+CT}} & \multicolumn{1}{c}{\text{GPb+CT}} \\
 & \multicolumn{1}{c}{\text{}} & \multicolumn{1}{c}{\text{}} & \multicolumn{1}{c}{\text{son}} & \multicolumn{1}{c}{\text{o1}} & \multicolumn{1}{c}{\text{son}} & \multicolumn{1}{c}{\text{o1}} & \multicolumn{1}{c}{\text{}} & \multicolumn{1}{c}{\text{}} & \multicolumn{1}{c}{\text{son}} & \multicolumn{1}{c}{\text{o1}} & \multicolumn{1}{c}{\text{son}} & \multicolumn{1}{c}{\text{o1}} \\
\midrule
Anaphora (1) & 94.7 & \multicolumn{1}{c}{\textbf{96.7}} & 93.3 & \multicolumn{1}{c}{\underline{95.3}} & \multicolumn{1}{c}{\textbf{96.7}} & \multicolumn{1}{c}{\textbf{96.7}} & \multicolumn{1}{c}{\textbf{96.0}} & 94.0 & \multicolumn{1}{c}{\underline{95.3}} & 94.7 & 94.7 & 94.7 \\
Arg. Str. (1) & 88.7 & 87.3 & \multicolumn{1}{c}{\textbf{90.7}} & \multicolumn{1}{c}{\underline{90.0}} & 88.0 & 84.0 & \multicolumn{1}{c}{\textbf{90.0}} & \multicolumn{1}{c}{\underline{89.3}} & \multicolumn{1}{c}{\underline{89.3}} & 88.0 & \multicolumn{1}{c}{\underline{89.3}} & 86.7 \\
Reflexives (1) & 94.7 & \multicolumn{1}{c}{\underline{96.7}} & \multicolumn{1}{c}{\textbf{100.0}} & \multicolumn{1}{c}{\textbf{100.0}} & \multicolumn{1}{c}{\textbf{100.0}} & \multicolumn{1}{c}{\textbf{100.0}} & \multicolumn{1}{c}{\underline{96.0}} & 93.3 & \multicolumn{1}{c}{\textbf{100.0}} & \multicolumn{1}{c}{\textbf{100.0}} & \multicolumn{1}{c}{\textbf{100.0}} & \multicolumn{1}{c}{\textbf{100.0}} \\
Subj.-Pred. Agr (5) & 94.1 & 95.9 & 95.6 & \multicolumn{1}{c}{\underline{96.5}} & \multicolumn{1}{c}{\textbf{98.1}} & 95.1 & 96.8 & 97.2 & 96.4 & 96.9 & \multicolumn{1}{c}{\underline{97.7}} & \multicolumn{1}{c}{\textbf{98.1}} \\
aspect (3) & 90.7 & 86.7 & \multicolumn{1}{c}{\textbf{100.0}} & 98.9 & \multicolumn{1}{c}{\textbf{100.0}} & \multicolumn{1}{c}{\underline{99.8}} & 96.2 & 92.2 & \multicolumn{1}{c}{\textbf{100.0}} & \multicolumn{1}{c}{\underline{99.3}} & \multicolumn{1}{c}{\textbf{100.0}} & \multicolumn{1}{c}{\underline{99.3}} \\
government (2) & 96.3 & 96.0 & \multicolumn{1}{c}{\underline{98.3}} & 97.0 & \multicolumn{1}{c}{\textbf{99.0}} & 98.0 & 97.7 & 97.3 & \multicolumn{1}{c}{\textbf{99.7}} & \multicolumn{1}{c}{\underline{99.0}} & \multicolumn{1}{c}{\underline{99.0}} & 98.7 \\
word formation (1) & 98.7 & \multicolumn{1}{c}{\textbf{100.0}} & \multicolumn{1}{c}{\textbf{100.0}} & \multicolumn{1}{c}{\textbf{100.0}} & \multicolumn{1}{c}{\textbf{100.0}} & \multicolumn{1}{c}{\underline{99.3}} & 98.7 & \multicolumn{1}{c}{\underline{99.3}} & \multicolumn{1}{c}{\underline{99.3}} & 98.7 & \multicolumn{1}{c}{\textbf{100.0}} & 97.3 \\
\midrule
Average & 94.0 & 94.2 & \multicolumn{1}{c}{\underline{96.8}} & 96.8 & \multicolumn{1}{c}{\textbf{97.4}} & 96.1 & 95.9 & 94.7 & \multicolumn{1}{c}{\underline{97.2}} & 96.7 & \multicolumn{1}{c}{\textbf{97.2}} & 96.4 \\
\bottomrule
\end{tabular}
\caption{Different prompting conditions for Russian-language RuBLiMP dataset}
\label{tab:RUBLIMP_main_table}
\end{table*}

\subsection{BLiMP Paradigms (English)}
\label{sec:blimp-analysis}
Table  \ref{tab:BLIMP_main_table} presents BLiMP results (for additional results using Llama please see Appendix \ref{sec:appendix_full_baselines}).

The grammar prompting conditions consistently outperforms basic and chain-of-thought conditions, with GPb+CoT(o1) performing best; for example, Sonnet and gpt-4o achieve strongest performance with GPb+CoT (92.7\% and 95.2\%), and Llama follows at 88.7\%. 

Sonnet's and o1's grammar prompts vary in effectiveness by phenomenon: o1's prompt underperforms Sonnet's for argument structure and binding, while excelling at filler-gap dependencies. Interestingly, in the GP condition, the smaller models perform better with Sonnet's explanations while larger models prefer o1's.

Grammar prompting shows dramatic improvements in some categories -- NPI licensing (c.f. the `popsicle' example from the introduction) improves from 73.3\% to 98.7\% for gpt-4o and reaches perfect accuracy with Sonnet, with similar gains in quantifier scope -- though some complex phenomena like binding and island effects show only modest improvements.


\subsection{SLING Paradigms (Chinese)}
\label{sec:sling_analysis}
Table \ref{tab:SLING_main_table} shows SLING results (for additional results using Llama please see Appendix \ref{sec:appendix_full_baselines}). Grammar prompting again yields substantial gains, with GPb+CoT optimal: gpt-3.5's average climbs from 74.0\% to 89.3\%, and Llama leaps from 82.4\% to 96.3\%. While smaller models start at 70--80\% and show large gains, Llama and larger models improve from 88--92\% to 95--100\% with GPb+CoT. Sonnet and o1's prompts perform similarly, though Sonnet's explanations are better for gpt-3.5 on alternative questions and definiteness paradigms. 

The alternative question paradigm, which tests the constraint that the adverb \textit{haishi} cannot co-occur with question particle \textit{ma}, shows a striking pattern: CoT alone leads to systematic errors in smaller models (GPT-3.5: 16.7\%, Haiku: 31.3\%, Llama: 8.5\%), but GPb and GPb+CoT achieve near-perfect results. 
Aspect particles improve from 70-90\% to the high 90s with GPb+CoT, while classifier-noun paradigms show minimal gains due to the long-distance dependency and lexical nature. Definiteness and polarity paradigms, which test simple adverbs and demonstratives, consistently reach near-perfect accuracy with GPb.

\subsection{RuBLiMP Paradigms (Russian)}
\label{sec:rublimp_analysis}
Table \ref{tab:RUBLIMP_main_table} shows RuBLiMP results (for additional results using Llama see Appendix \ref{sec:appendix_full_baselines}). Grammar prompting again shows consistent gains, with GPb+CoT optimal: gpt-4o and Sonnet again achieve the strongest results (94.8\% and 95.2\% with GPb+CoT), followed by Llama (93.7\%), with 5-15 point improvements over baseline across models. Prompt effectiveness varies by model and paradigm: smaller models struggle with Sonnet's argument structure and o1's government prompts, while Haiku strongly prefers Sonnet's aspect prompts (96.3\% vs o1's 88.7\%).

The reflexive possessor paradigm shows dramatic improvement, reaching near 100\% with GPb as models grasp the constraint that \textit{u + sebya} is allowed only as a locative and not a possessive. Aspect-related paradigms, which involve selecting perfect or imperfect verb form, improve from poor/moderate baselines to near-perfect scores with GPb+CoT, suggesting models know the morphology but need usage clarification. Case government improves from 70-90\% to 80-97\%, with Sonnet reaching 97-99\% in GPb+CoT.

\subsection{Three-shot Results} \label{sec:three-shot}
Table~\ref{tab:model_performance_comparison} reports a three-shot results alongside grammar prompting, contrasting \emph{pattern-matching from examples} with our rule-based GP approach.  With three labeled minimal pairs, many paradigms become solvable via surface similarity alone, offering a useful foil for testing formal grammatical competence.

The results diverge sharply by model size.  For the two LLMs (Sonnet and GPT-4o), 3-shot pattern matching is highly effective, outperforming GP on English and matching it on Chinese and Russian.  For the three SLMs (gpt-3.5, Haiku, and Llama), the picture reverses dramatically:
\emph{grammar prompting exceeds 3-sh by an average of 12.1-13.1 percentage points (pp)}.
GP therefore narrows the LLM--SLM performance gap precisely where few-shot learning leaves smaller models behind, reinforcing its value as a low-cost equalizer.


\begin{table}[!t]
\centering
\small
\begin{tabular}{@{}l l S[table-format=2.2] S[table-format=2.2] S[table-format=2.2] S[table-format=2.2]@{}}
\toprule
 & Setting & \multicolumn{1}{c}{Sonnet} & \multicolumn{1}{c}{GPT-4o} & \multicolumn{1}{c}{Haiku} & \multicolumn{1}{c}{GPT-3.5} \\
\midrule
\multirow{2}{*}{EN} 
  & 3-shot     & \textbf{97.31} & \textbf{94.48} & 73.63 & 65.26 \\
  & GP(b)-o1   & 90.17 & 87.86 & \textbf{82.33} & \textbf{77.92} \\
\noalign{\smallskip}\hdashline \noalign{\smallskip}
\multirow{2}{*}{CH} 
  & 3-shot     & 97.44 & 98.03 & 79.90 & 79.75 \\
  & GP(b)-o1   & \textbf{98.06} & \textbf{98.12} & \textbf{96.34} & \textbf{94.10} \\
\noalign{\smallskip}\hdashline \noalign{\smallskip}
\multirow{2}{*}{RU} 
  & 3-shot     & \textbf{97.37} & 96.61 & 83.21 & 75.59 \\
  & GP(b)-o1   & 97.19 & \textbf{97.36} & \textbf{94.36} & \textbf{87.75} \\
\bottomrule
\end{tabular}
\caption{Macroaverage performance of three-shot vs grammar prompting. Three-shot beats grammar prompting for large models in English, but grammar prompting is dramatically more effective \textit{on smaller LLMs} across all three languages.}
\label{tab:model_performance_comparison}
\end{table}

\subsection{Effect on the LLM--SLM Performance Gap}
\label{sec:slm-llm-gap}

To assess whether grammar prompting narrows the disparity between model
sizes, we group the five models into SLMs (gpt-3.5, Haiku, Llama) and LLMs (gpt-4o, Sonnet).  For each language we compute the
mean accuracy within each group and define the \textit{gap} as 
$\overline{\text{LLM}}-\overline{\text{SLM}}$.
The bottom block of Table~\ref{tab:slm_llm_gaps} reports the
cross-language average of these three per-language gaps.

In the zero-shot baseline, the mean LLM--SLM gap is
\(13.0\) pp.
Adding grammar prompting alone reduces the gap to \(10.4\) pp, a
\(20\%\) relative contraction.
When chain-of-thought reasoning is supplied in addition to the grammar
prompt (GP + CoT), the gap shrinks to \(5.8\) pp, a \(56\%\) reduction
relative to the baseline.
Hence a single LLM-generated rule explanation, especially when paired
with step-by-step reasoning, brings three SLMs to within roughly six
percentage points of gpt-4-class models across English, Chinese, and
Russian, without any additional training or task-specific tuning.

\begin{table}[!t]
\centering\small
\begin{tabular}{llccc}
\toprule
 & \textbf{Condition} & \textbf{SLM Avg} & \textbf{LLM Avg} & \textbf{Gap} \\
\midrule
\multirow{4}{*}{\textbf{English}}
 & Base      & 66.0 & 78.9 & 12.9 \\
 & CT        & 69.0 & 82.6 & 13.6 \\
 & GP        & 73.6 & 91.0 & 17.4 \\
 & GP+CT     & \textbf{84.5} & \textbf{94.0} &  \textbf{9.4} \\
\midrule
\multirow{4}{*}{\textbf{Chinese}}
 & Base      & 77.4 & 90.2 & 12.7 \\
 & CT        & 77.1 & 95.1 & 18.0 \\
 & GP        & 92.8 & 97.7 &  4.9 \\
 & GP+CT     & \textbf{94.9} & \textbf{98.3} &  \textbf{3.4} \\
\midrule
\multirow{4}{*}{\textbf{Russian}}
 & Base      & 81.5 & 95.0 & 13.4 \\
 & CT        & 82.4 & 94.5 & 12.1 \\
 & GP        & 87.9 & \textbf{96.8} &  8.9 \\
 & GP+CT     & \textbf{91.8} & 96.3 &  \textbf{4.4} \\
\midrule
\multirow{4}{*}{\textbf{Average}}
 & Base      & 75.0 & 88.0 & 13.0 \\
 & CT        & 76.2 & 90.7 & 14.5 \\
 & GP        & 84.8 & 95.2 & 10.4 \\
 & GP+CT     & \textbf{90.4} & \textbf{96.2} &  \textbf{5.8} \\
\bottomrule
\end{tabular}
\caption{Accuracy (\%) of small language models (SLMs) vs.\ large language models (LLMs) under different prompting conditions. “Gap” is the difference LLM-SLM; lower is better for SLMs. The bottom block reports the cross-language average.}
\label{tab:slm_llm_gaps}
\end{table}

\section{Discussion}
\label{sec:discussion}

Our results across English, Chinese and Russian reveal clear patterns how grammar prompting works for different linguistic phenomena. It is most effective for phenomena governed by simple distributional constraints and categorical distinctions, such as negative polarity items in English, aspect marking in Russian and Chinese, or alternative questions in Chinese. These often reach near-perfect accuracy under grammar prompting, likely because they involve specific functional elements (e.g., \textit{any}, \textit{haishi}, \textit{sebya}) that operate under specific licensing environments or paradigmatic choices like Russian perfect/imperfect verb aspect. {This accuracy jump is especially pronounced for SLMs.}

In contrast, phenomena that demand fine-grained constituent recognition (e.g.\ English island constraints) or language-specific lexical knowledge (e.g.\ Chinese classifier-noun agreement, wh-fronting) show smaller but still consistent accuracy gains. Appendix~\ref{sec:appendix_success_and_failure} documents instances of successes and characteristic failure modes: when the model \emph{correctly identifies} the relevant constituent, grammar prompting enables rule-based judgments even on island violations; when it misparses or loses track of its own reasoning (as GPT-3.5 occasionally does on Chinese wh-fronting), the benefit is muted.  These errors appear to stem less from the prompting paradigm itself than from (i) occasional constituent identification failures, (ii) the added cognitive load of explicit chain-of-thought reasoning, and (iii) our deliberate decision to omit illustrative examples.  Addressing any of these factors -- better parsing, better reasoning models, or simply adding in-context examples -- could therefore push performance on the ``hard'' paradigms substantially higher.


We opened this paper with an incorrect conclusion drawn by Sonnet using CoT. 

\begin{tcolorbox}[
  colframe=black, colback=purple!10, 
  boxrule=0.8pt, sharp corners,  
  width=\columnwidth, 
]
\begin{itemize}
  \item[(A)] Only a popsicle that Danielle admires ever freezes. \checkmark
  \item[(B)] A popsicle that only Danielle admires ever freezes. $\times$
\end{itemize}
\end{tcolorbox}

With grammar prompting, Sonnet now reasons:

\textit{``The key difference is the placement of "only". In Sentence A, it's correctly placed at the beginning of the entire noun phrase, scoping over `a popsicle that Danielle admires'. In Sentence B, it's incorrectly placed inside the relative clause, which changes the meaning and doesn't follow the grammar rule explained. Therefore, Sentence A is more grammatically acceptable according to the provided explanation.''}

In the introduction, we observed how models can err by abstracting away surface but essential linguistic form in the attempt to decipher communicative intent; now, we see how grammar prompting reduces problematic `semantically-driven reasoning' by 1) harnessing the model's metalinguistic ability to generate a grammar prompt and 2) feeding the grammar prompt back to the model, thereby directing its reasoning process to operate over the relevant linguistic structure.
\section{Conclusion}
LLMs are trained to distill away surface forms to capture meaning. This `meaning-first' strategy powers functional tasks such as translation but can sabotage syntactic reasoning. Grammar prompting recalibrates that bias and does so \emph{as a low-cost equalizer between model sizes}.  With a single rule explanation generated by a large LLM and fed back with chain-of-thought, three SLMs finish on average within 5.8 pp of GPT-4-class models across English, Chinese, and Russian, cutting the capacity gap by 56\%.

\emph{From general to specific reasoning.} Grammar prompting anchors the model's attention on relevant grammatical categories and constraints.

\emph{From implicit to explicit knowledge.}  
It turns latent metalinguistic knowledge into an explicit rule the model can apply, boosting accuracy even when data-driven cues are sparse.

The explain-then-process paradigm is attractive for (i) low-resource languages, where training data are scarce, and (ii) deployments that must rely on smaller, cheaper models when frontier LLMs are impractical.

\section*{Acknowledgments}
We are grateful to the anonymous reviewers and the area chair for their insightful and constructive feedback. 

\section*{Limitations}
Our study has several important limitations. First, we only elicited grammar prompts from large models (Sonnet and o1), not exploring whether smaller models could generate effective explanations. Additionally, while we used English for all prompts, recent work suggests that prompting in the non-dominant language might yield better results in some cases \cite{behzad-etal-2024-ask}.

Second, our decision to exclude example sentences from grammar prompts, while methodologically sound for isolating the effect of explanations, may have also limited the prompts' potential effectiveness, particularly for complex phenomena like island constraints.

Finally, we focused solely on high-resource languages to control for vocabulary familiarity, but it remains untested how effective grammar prompting would be for low-resource languages: while low-resource languages often have well-documented grammatical systems that could inform prompt generation, their limited representation in LLM training data might affect both the models' ability to generate accurate grammatical explanations and to apply them correctly. This limitation warrants future investigation. 

\bibliography{custom}


\appendix

\section{Appendix: Instruction Template and Grammar Prompt Examples}
\label{sec:instructiontemplate}

Figure~\ref{app:1}, ~\ref{app:2}, and \ref{app:3} contain the full instruction template and three generated prompts: an `expert-oriented prompt' by Sonnet, a `beginner-oriented prompt' by Sonnet, and a `beginner-oriented prompt' by o1, respectively.

\begin{figure*}[!t]
\begin{tcolorbox}[
  colframe=black, colback=white, 
  boxrule=0.8pt, sharp corners, 
  width=\textwidth
]
\small 

\textbf{\underline{Instruction} for `left branch island echo question' BLiMP paradigm:}

\textbf{Task:} Explain, to a novice learner, `left branch island echo question' in English (no need to use this term).

\textbf{Context:} Help a novice learner understand and identify correct usage of this 'left branch island echo question' concept to pass a test.

\textbf{Requirements:}
\begin{itemize}
    \item Clear, non-technical title
    \item Explain key rules and patterns
    \item Include specific grammatical details
    \item Provide tips for distinguishing correct and incorrect usage
    \item Use appropriate English words freely
    \item Avoid full example sentences
    \item Tailor complexity and language to a novice learner
\end{itemize}

\textbf{Examples (for reference, do not include):}

Good sentence: Many boys will embarrass what patient?\\
Bad sentence: What will many boys embarrass patient?

Good sentence: Sara was insulting what student?\\
Bad sentence: What was Sara insulting student?

Good sentence: Gina is helping which adults?\\
Bad sentence: Which is Gina helping adults?

\textbf{Target length:} about 250 words

\end{tcolorbox}
\caption{Instruction for generating grammar explanations.}\label{app:1} \vspace{-0.4cm}
\end{figure*}


\begin{figure*}[!t]
\begin{tcolorbox}[
  colframe=black, colback=white, 
  boxrule=0.8pt, sharp corners, 
  width=\textwidth
]
\small 

\textbf{\underline{Sonnet-generated Expert grammar prompt}}
\vspace{.5em}

\textbf{Title: Interrogative Phrase Extraction in Echo Questions}

For expert linguists, it's crucial to understand the constraints on extracting interrogative phrases from certain syntactic positions in echo questions. This phenomenon relates to the left periphery of noun phrases and the limitations on moving wh-elements.

In grammatical constructions, the interrogative element (e.g., "what," "which") can be extracted when it's the entire object of a verb. However, when the interrogative element is embedded within a larger noun phrase, particularly as a determiner or modifier on the left edge, extraction becomes ungrammatical.

Key syntactic rules to note:
\begin{enumerate}
    \item The interrogative element must remain in situ when it's part of a complex noun phrase.
\item Echo questions preserve the basic word order of the original statement.
\item The wh-word replaces only the specific element being questioned, not the entire phrase.
\end{enumerate}

Linguists should pay attention to the following grammatical details:
\begin{itemize}
    \item Determiner position within noun phrases
\item  Argument structure of verbs
\item  Prosodic patterns typical of echo questions
\end{itemize}

To distinguish correct usage, consider:
\begin{itemize}
\item  Is the wh-word modifying a noun or standing alone as an argument?
\item  Does the question maintain the original sentence structure?
\item  Is there any violation of island constraints?
\end{itemize}

When analyzing potential violations, focus on the internal structure of noun phrases and the position of the interrogative element within them. Correct usage will maintain the integrity of the noun phrase, while incorrect usage will attempt to extract the wh-element from its embedded position.
\end{tcolorbox}
\caption{Sonnet-generated Expert grammar prompt for `left branch island echo question' in the BLiMP paradigm.}\vspace{-0.5cm}\label{app:2}
\end{figure*}

\begin{figure*}[!t]
\begin{tcolorbox}[
  colframe=black, colback=white, 
  boxrule=0.8pt, sharp corners, 
  width=\textwidth
]
\small 


\textbf{\underline{Sonnet-generated Beginner grammar prompt}}
\vspace{.5em}

\textbf{Title: Understanding Special Question Patterns in English}

In English, there's a particular way of asking questions that might seem tricky at first. This pattern involves asking about a specific part of a sentence while keeping most of the original sentence structure intact.

Key Rules:
\begin{enumerate}
\item The question word (like "what" or "which") stays at the end of the question.
\item The rest of the sentence remains mostly unchanged from its statement form.
\item This pattern is often used to express surprise or to confirm information.
\end{enumerate}

Grammatical Details:
\begin{itemize}
\item These questions typically use "what," "which," or "whose" as the question word.
\item The main verb and its tense remain the same as in the original statement.
\item The subject and any other elements stay in their original positions.
\end{itemize}

Tips for Correct Usage:
\begin{itemize}
\item Imagine the statement first, then add the question word at the end.
\item Don't move the question word to the beginning of the sentence.
\item Keep all other words in their original order.
\item Make sure the question word matches what you're asking about (e.g., "what" for objects, "which" for choices).
\end{itemize}

To spot incorrect usage:
\begin{itemize}
\item Look out for questions where the question word is at the beginning instead of the end.
\item Be wary of any rearrangement of words from the original statement form.
\item Check if any words are missing that would be present in a statement.
\end{itemize}

Remember, this question form is used in specific contexts, often to express surprise or seek confirmation. It's not the standard way to form all questions in English, but it's important to recognize and use correctly when appropriate.

\vspace{2 em}
\textbf{\underline{o1-generated Beginner grammar prompt}}\\
\vspace{.5em}
\textbf{A Simple Way to Ask Follow-Up Questions}

Sometimes you need to clarify or confirm information without changing the basic sentence too much. In these follow-up questions, the word you use (like “what” or “which”) stays close to the place in the sentence where the missing detail occurs, instead of leaping to the beginning. Imagine a statement such as “X is embarrassing Y” or “A was doing B with C.” If you want to ask about the missing piece, replace it with “what” or “which” and keep the rest of the sentence order mostly the same (e.g., “A was doing B with which…”).
\vspace{.5em}
One useful tip is to look at how you would normally say the statement. Then, swap the unknown element for the question word, placing it near the end. This helps keep the flow. If you force the question word (“what,” “which,” and so on) to the front of the sentence, it may create an awkward or incorrect structure in these follow-up situations.
\vspace{.5em}
Another guideline is to ask yourself: “Am I simply confirming or clarifying something?” If yes, keep the question word inside or at the end of the sentence. If something sounds unnatural or is missing key parts (for example, if you have no place for the question word to fit smoothly), that likely indicates a mistake.
\vspace{.5em}
Finally, practice by taking short statements and gently turning them into questions. Ask: “Whose detail am I missing?” Then logically insert your question word. This approach will help ensure you keep the sentence structure coherent and easily understandable.
\end{tcolorbox}
\caption{Sonnet and o1-generated Beginner grammar prompt for `left branch island echo question' in the BLiMP paradigm.}\vspace{-0.4cm}\label{app:3}
\end{figure*}

\section{Appendix: Choosing Challenging Paradigms to Test}
\label{sec:appendix_full_baselines}
Initially, we tested gpt-4o to ascertain which paradigms its performance is low enough to allow for improvement over the base condition. 
Tables~\ref{tab:comparison}, \ref{tab:comparison-sling}, and \ref{tab:comparison-rublimp} show these results for BLiMP, SLING, and RuBLiMP datasets, respectively. Table \ref{tab:big-llama-summary} presents the results of Llama models.

\begin{table*}[!t]
\centering
\small
\begin{tabular}{llcc}
\toprule
Category & Paradigm & gpt-3.5 & gpt-4o \\
\midrule
\multirow{7}{*}{\textsc{NPI licensing}} & matrix\_question\_npi\_licensor\_present & 93.3 & \textbf{98.0} \\
 & npi\_present\_1 & 88.7 & \textbf{96.7} \\
 & npi\_present\_2 & 96.0 & \textbf{100.0} \\
 & only\_npi\_licensor\_present & 96.0 & \textbf{98.7} \\
 & \textbf{only\_npi\_scope} & \textbf{53.3} & 52.0 \\
 & sentential\_negation\_npi\_licensor\_present & 91.3 & \textbf{99.3} \\
 & \textbf{sentential\_negation\_npi\_scope} & 70.7 & \textbf{89.3} \\
\multirow{2}{*}{\textsc{anaphora agreement}} & anaphor\_gender\_agreement & \textbf{100.0} & 100.0 \\
 & anaphor\_number\_agreement & 99.3 & \textbf{100.0} \\
\multirow{9}{*}{\textsc{argument structure}} & animate\_subject\_passive & 82.7 & \textbf{92.0} \\
 & animate\_subject\_trans & 87.3 & \textbf{92.7} \\
 & causative & 86.0 & \textbf{91.3} \\
 & \textbf{drop\_argument} & 67.3 & \textbf{89.3} \\
 & inchoative & 85.3 & \textbf{90.7} \\
 & intransitive & 77.3 & \textbf{97.3} \\
 & passive\_1 & 90.7 & \textbf{95.3} \\
 & passive\_2 & 94.0 & \textbf{94.7} \\
 & transitive & 84.0 & \textbf{97.3} \\
\multirow{7}{*}{\textsc{binding}} & principle\_A\_c\_command & 91.3 & \textbf{98.7} \\
 & principle\_A\_case\_1 & \textbf{100.0} & 100.0 \\
 & principle\_A\_case\_2 & \textbf{98.0} & 96.7 \\
 & principle\_A\_domain\_1 & 98.0 & \textbf{99.3} \\
 & principle\_A\_domain\_2 & 88.7 & \textbf{94.0} \\
 & \textbf{principle\_A\_domain\_3} & 74.7 & \textbf{81.3} \\
 & \textbf{principle\_A\_reconstruction} & 43.3 & \textbf{63.3} \\
\multirow{5}{*}{\textsc{control/raising}} & \textbf{existential\_there\_object\_raising} & \textbf{88.0} & 86.0 \\
 & existential\_there\_subject\_raising & 90.7 & \textbf{92.7} \\
 & \textbf{expletive\_it\_object\_raising} & 82.7 & \textbf{88.7} \\
 & \textbf{tough\_vs\_raising\_1} & 72.0 & \textbf{77.3} \\
 & tough\_vs\_raising\_2 & 76.0 & \textbf{96.7} \\
\multirow{8}{*}{\textsc{determiner-noun agreement}} & determiner\_noun\_agreement\_1 & 98.7 & \textbf{100.0} \\
 & determiner\_noun\_agreement\_2 & 99.3 & \textbf{100.0} \\
 & determiner\_noun\_agreement\_irregular\_1 & 90.7 & \textbf{100.0} \\
 & determiner\_noun\_agreement\_irregular\_2 & 97.3 & \textbf{100.0} \\
 & determiner\_noun\_agreement\_with\_adj\_2 & 94.0 & \textbf{96.7} \\
 & determiner\_noun\_agreement\_with\_adj\_irregular\_1 & 88.7 & \textbf{100.0} \\
 & determiner\_noun\_agreement\_with\_adj\_irregular\_2 & 96.7 & \textbf{98.0} \\
 & determiner\_noun\_agreement\_with\_adjective\_1 & 94.0 & \textbf{99.3} \\
\multirow{2}{*}{\textsc{distractor agreement}} & distractor\_agreement\_relational\_noun & \textbf{100.0} & 100.0 \\
 & distractor\_agreement\_relative\_clause & 93.3 & \textbf{94.0} \\
\multirow{2}{*}{\textsc{ellipsis}} & \textbf{ellipsis\_n\_bar\_1} & 80.0 & \textbf{82.0} \\
 & \textbf{ellipsis\_n\_bar\_2} & \textbf{64.0} & 55.3 \\
\multirow{7}{*}{\textsc{filler-gap}} & wh\_questions\_object\_gap & 86.7 & \textbf{98.0} \\
 & wh\_questions\_subject\_gap & 81.3 & \textbf{97.3} \\
 & \textbf{wh\_questions\_subject\_gap\_long\_distance} & 66.7 & \textbf{90.7} \\
 & wh\_vs\_that\_no\_gap & 78.0 & \textbf{98.0} \\
 & wh\_vs\_that\_no\_gap\_long\_distance & 68.0 & \textbf{96.7} \\
 & \textbf{wh\_vs\_that\_with\_gap} & 50.7 & \textbf{81.3} \\
 & \textbf{wh\_vs\_that\_with\_gap\_long\_distance} & 48.0 & \textbf{54.0} \\
\multirow{2}{*}{\textsc{irregular forms}} & irregular\_past\_participle\_adjectives & 99.3 & \textbf{100.0} \\
 & irregular\_past\_participle\_verbs & \textbf{100.0} & 100.0 \\
\multirow{8}{*}{\textsc{island effect}} & \textbf{adjunct\_island} & \textbf{86.7} & 86.7 \\
 & \textbf{complex\_NP\_island} & 63.3 & \textbf{68.0} \\
 & coordinate\_structure\_constraint\_complex\_left\_branch & 84.0 & \textbf{93.3} \\
 & coordinate\_structure\_constraint\_object\_extraction & 90.7 & \textbf{92.0} \\
 & \textbf{left\_branch\_island\_echo\_question} & 78.0 & \textbf{92.0} \\
 & left\_branch\_island\_simple\_question & 97.3 & \textbf{99.3} \\
 & \textbf{sentential\_subject\_island} & \textbf{48.0} & 48.0 \\
 & \textbf{wh\_island} & 66.7 & \textbf{86.7} \\
\multirow{4}{*}{\textsc{quantifiers}} & existential\_there\_quantifiers\_1 & 94.7 & \textbf{97.3} \\
 & \textbf{existential\_there\_quantifiers\_2} & 54.7 & \textbf{79.3} \\
 & superlative\_quantifiers\_1 & 88.7 & \textbf{95.3} \\
 & \textbf{superlative\_quantifiers\_2} & 54.7 & \textbf{79.3} \\
\multirow{4}{*}{\textsc{subject-verb agreement}} & irregular\_plural\_subject\_verb\_agreement\_1 & \textbf{100.0} & 100.0 \\
 & irregular\_plural\_subject\_verb\_agreement\_2 & \textbf{100.0} & 100.0 \\
 & regular\_plural\_subject\_verb\_agreement\_1 & 98.0 & \textbf{100.0} \\
 & regular\_plural\_subject\_verb\_agreement\_2 & 98.7 & \textbf{100.0} \\
\midrule
 & Average & 84.0 & \underline{91.2} \\
\bottomrule
\end{tabular}
\caption{All 67 BLiMP paradigm results}
\label{tab:comparison}
\end{table*}

\begin{table*}[!t]
\centering
\small
\vspace{0.2cm}
\begin{tabular}{llc}
\toprule
Category & Paradigm & gpt-4o-2024-08-06 accuracy \\
\midrule
\multirow{1}{*}{\textsc{Alternative Question}} & \textbf{haishi\_ma} & 92.6 \\
\multirow{7}{*}{\textsc{Classifier-Noun}} & \textbf{cl\_adj\_comp\_noun} & 68.6 \\
 & \textbf{cl\_adj\_comp\_noun\_v2} & 99.3 \\
 & \textbf{cl\_adj\_simple\_noun} & 95.3 \\
 & \textbf{cl\_comp\_noun} & 72.0 \\
 & \textbf{cl\_comp\_noun\_v2} & 98.0 \\
 & \textbf{cl\_simple\_noun} & 95.3 \\
 & \textbf{dem\_cl\_swap} & 100.0 \\
\multirow{2}{*}{\textsc{Definiteness Effect}} & \textbf{definiteness\_demonstrative} & 88.0 \\
 & \textbf{definiteness\_every} & 96.0 \\
\multirow{3}{*}{\textsc{Polarity Item}} & \textbf{any} & 85.3 \\
 & \textbf{even\_wh} & 98.0 \\
 & \textbf{more\_or\_less} & 85.3 \\
\multirow{2}{*}{\textsc{Wh Fronting}} & \textbf{bare\_wh} & 100.0 \\
 & \textbf{mod\_wh} & 86.6 \\
\multirow{4}{*}{\textsc{Aspect}} & \textbf{temporal\_guo} & 100.0 \\
 & \textbf{temporal\_le} & 100.0 \\
 & \textbf{zai\_guo} & 96.6 \\
 & \textbf{zai\_le} & 77.3 \\
 & \textbf{zai\_no\_le} & 96.6 \\
\midrule
 & Average & \underline{92.8} \\
\bottomrule
\end{tabular}
\caption{SLING Paradigm Results}
\label{tab:comparison-sling}
\end{table*}

\begin{table*}[!t]
\centering
\small
\begin{tabular}{llc}
\toprule
Category & Paradigm & gpt-4o-2024-08-06 accuracy \\
\midrule
\multirow{2}{*}{\textsc{Anaphor Agreement}} & \textbf{anaphor\_agreement\_gender} & 92.0 \\
 & anaphor\_agreement\_number & 98.0 \\
\multirow{5}{*}{\textsc{Argument Structure}} & transitive\_verb & 100.0 \\
 & transitive\_verb\_iobject & 99.3 \\
 & transitive\_verb\_object & 99.3 \\
 & \textbf{transitive\_verb\_passive} & 88.7 \\
 & transitive\_verb\_subject & 96.0 \\
\multirow{3}{*}{\textsc{Aspect}} & \textbf{change\_duration\_aspect} & 95.3 \\
 & \textbf{change\_repetition\_aspect} & 96.0 \\
 & \textbf{deontic\_imperative\_aspect} & 84.7 \\
\multirow{5}{*}{\textsc{Government}} & adposition\_government & 100.0 \\
 & nominalization\_case & 96.0 \\
 & \textbf{verb\_acc\_object} & 96.7 \\
 & \textbf{verb\_gen\_object} & 94.7 \\
 & verb\_ins\_object & 97.3 \\
\multirow{3}{*}{\textsc{Negation}} & indefinite\_pronoun\_to\_negative & 98.0 \\
 & negative\_concord & 98.7 \\
 & negative\_pronoun\_to\_indefinite & 98.0 \\
\multirow{6}{*}{\textsc{Noun Phrase Agreement}} & floating\_quantifier\_agreement\_case & 100.0 \\
 & floating\_quantifier\_agreement\_gender & 98.7 \\
 & floating\_quantifier\_agreement\_number & 97.3 \\
 & np\_agreement\_case & 97.3 \\
 & np\_agreement\_gender & 99.3 \\
 & np\_agreement\_number & 99.3 \\
\multirow{1}{*}{\textsc{Reflexives}} & \textbf{external\_possessor} & 95.3 \\
\multirow{11}{*}{\textsc{Subject-Predicate Agreement}} & clause\_subj\_predicate\_agreement\_gender & 98.7 \\
 & clause\_subj\_predicate\_agreement\_number & 94.7 \\
 & clause\_subj\_predicate\_agreement\_person & 96.7 \\
 & \textbf{genitive\_subj\_predicate\_agreement\_gender} & 95.3 \\
 & genitive\_subj\_predicate\_agreement\_number & 98.7 \\
 & genitive\_subj\_predicate\_agreement\_person & 98.0 \\
 & \textbf{noun\_subj\_predicate\_agreement\_gender} & 90.0 \\
 & noun\_subj\_predicate\_agreement\_number & 100.0 \\
 & noun\_subj\_predicate\_agreement\_person & 98.7 \\
 & \textbf{subj\_predicate\_agreement\_gender\_attractor} & 91.3 \\
 & \textbf{subj\_predicate\_agreement\_number\_attractor} & 96.7 \\
\multirow{3}{*}{\textsc{Tense and Mood}} & conj\_verb\_tense & 99.3 \\
 & single\_verb\_tense & 99.3 \\
 & tense\_marker & 97.3 \\
\multirow{6}{*}{\textsc{Word Formation}} & \textbf{add\_new\_suffix} & 96.7 \\
 & add\_verb\_prefix & 98.7 \\
 & change\_declension\_ending & 98.7 \\
 & change\_declension\_ending\_has\_dep & 98.7 \\
 & change\_verb\_conjugation & 98.0 \\
 & change\_verb\_prefixes\_order & 100.0 \\
\midrule
 & Average & \underline{96.9} \\
\bottomrule
\end{tabular}
\caption{All 45 RuBLiMP paradigm results}
\label{tab:comparison-rublimp}
\end{table*}

\begin{table*}[!t]
\centering
\setlength{\tabcolsep}{14pt}
\small
\begin{tabular}{p{3cm} S[table-format=2.1] S[table-format=2.1] S[table-format=2.1] S[table-format=2.1]}
\toprule
Categories & \multicolumn{1}{c}{Base} & \multicolumn{1}{c}{CT} & \multicolumn{1}{c}{GPb} & \multicolumn{1}{c}{GPb+CT} \\
\midrule 
\multicolumn{5}{c}{\cellcolor{gray!20}\textsc{English \textbf{(BLIMP)}}} \\
NPI lic. (2) & 62.3 & 66.7 & 87.7 & \multicolumn{1}{c}{\textbf{98.3}} \\
Arg. Str. (1) & 80.7 & \multicolumn{1}{c}{\textbf{82.6}} & 70.7 & 76.4 \\
binding (2) & 87.7 & 81.0 & 85.3 & \multicolumn{1}{c}{\textbf{90.3}} \\
Cntrl/Rs. (3) & 85.3 & 88.0 & 87.6 & \multicolumn{1}{c}{\textbf{91.3}} \\
ellipsis (2) & 54.0 & 70.3 & 69.3 & \multicolumn{1}{c}{\textbf{87.5}} \\
filler gap (3) & 55.7 & 61.3 & 94.4 & \multicolumn{1}{c}{\textbf{97.6}} \\
island ef. (5) & 65.1 & 70.8 & 71.9 & \multicolumn{1}{c}{\textbf{76.5}} \\
quantifiers (2) & 60.3 & 57.1 & 81.0 & \multicolumn{1}{c}{\textbf{91.3}} \\
\noalign{\smallskip}\hdashline \noalign{\smallskip}
Average (BLiMP) & 68.9 & 72.2 & 81.0 & \multicolumn{1}{c}{\textbf{88.7}} \\

\multicolumn{5}{c}{\cellcolor{gray!20}\textsc{Chinese \textbf{(SLING)}}} \\

Alt. Question (1) & 40.7 & 8.5 & \multicolumn{1}{c}{\textbf{100.0}} & \multicolumn{1}{c}{\textbf{100.0}} \\
Aspect (4) & 83.7 & 88.4 & 95.0 & \multicolumn{1}{c}{\textbf{97.0}} \\
Classifier-Noun (6) & 87.3 & 87.0 & 83.0 & \multicolumn{1}{c}{\textbf{88.1}} \\
Definit. Ef. (3) & 92.2 & 94.1 & \multicolumn{1}{c}{\textbf{100.0}} & \multicolumn{1}{c}{\textbf{100.0}} \\
Polarity Item (3) & 92.9 & 92.4 & \multicolumn{1}{c}{\textbf{100.0}} & \multicolumn{1}{c}{\textbf{100.0}} \\
Wh Fronting (2) & 97.7 & 97.0 & \multicolumn{1}{c}{\textbf{100.0}} & \multicolumn{1}{c}{\textbf{100.0}} \\
\noalign{\smallskip}\hdashline \noalign{\smallskip}
Average (SLING) & 82.4 & 77.9 & 96.3 & \multicolumn{1}{c}{\textbf{97.5}} \\

\multicolumn{5}{c}{\cellcolor{gray!20}\textsc{Russian \textbf{(RUBLIMP)}}} \\

Anaphora (1) & 92.7 & 93.3 & 87.3 & \multicolumn{1}{c}{\textbf{95.3}} \\
Arg. Str. (1) & 86.7 & \multicolumn{1}{c}{\textbf{90.5}} & 87.3 & 90.3 \\
Reflexives (1) & 88.0 & 96.0 & \multicolumn{1}{c}{\textbf{100.0}} & \multicolumn{1}{c}{\textbf{100.0}} \\
Subj.-Pred. Agr (5) & 91.9 & \multicolumn{1}{c}{\textbf{94.0}} & 89.5 & 92.0 \\
aspect (3) & 75.6 & 79.9 & 91.1 & \multicolumn{1}{c}{\textbf{95.1}} \\
government (2) & 83.7 & 86.0 & 88.0 & \multicolumn{1}{c}{\textbf{94.9}} \\
word formation (1) & 92.0 & \multicolumn{1}{c}{\textbf{95.3}} & 87.3 & 87.8 \\
\noalign{\smallskip}\hdashline \noalign{\smallskip}
Average (RuBLiMP) & 87.2 & 90.7 & 90.1 & \multicolumn{1}{c}{\underline{93.6}} \\
\midrule
Average over all & 78.8 & 80.0 & 88.4 & \multicolumn{1}{c}{\underline{92.8}} \\
\bottomrule
\end{tabular}
\caption{Llama results on BLiMP, SLING and RuBLiMP over basic, CT, GPb(o1) and GPb+CT(o1)}
\label{tab:big-llama-summary}
\end{table*}

\label{sec:appendix_control}

\section{Appendix: Success and Failure Cases}
\label{sec:appendix_success_and_failure}

We present examples of models' success and failure cases under two grammatical constraints: the sentential island constraint in English and the wh-fronting constraint in Chinese. These cases demonstrate the strengths of grammar prompting over a solely-CoT approach, and show where the models still fail to fully utilize grammar explanations.

\subsection{English Sentential Island Constraint (Sonnet)}%
We asked Claude 3.5 Sonnet to judge which of the following sentences is more grammatically acceptable:

\begin{tcolorbox}[
  colframe=black, colback=purple!10,
  boxrule=0.8pt, sharp corners,
  width=\columnwidth,
]
\begin{itemize}[leftmargin=1em, nosep]
  \item[(A)] Who had the patients' cleaning those banks upset.
  \item[(B)] Who had the patients' cleaning upset those banks.
\end{itemize}
\end{tcolorbox}

\textbf{(i) CoT in basic prompt (CoT failure case)}\\
\textit{Excerpt from Sonnet’s response:}
\begin{quote}
\small
“\ldots Both sentences are confusing\ldots In Sentence B: \ldots `Who' is the subject, `had' is the main verb, `the patients' cleaning' is a noun phrase (object), \ldots Sentence B's structure is closer to a correct English sentence pattern. \ldots Sentence A's structure is less common and more difficult to interpret correctly \ldots.”
\end{quote}

\textit{Analysis:} Sonnet fails to identify the relevant grammatical constituents. It evaluates based on surface fluency and frequency, lacking access to the rule that governs island violations.

\vspace{0.5em}
\textbf{(ii) GP-CoT(o1) prompt (GP failure case)}\\
\textit{Excerpt from Sonnet’s response:}
\begin{quote}
“1. First, we need to identify the main subject \ldots In both sentences, \ldots we can see that the subject is complex \ldots: `the patients' cleaning.' This acts like the `protective bubble' mentioned in the explanation. The explanation states that we shouldn't move words from inside this subject clause when forming questions\ldots Looking at Sentence A\ldots this sentence has moved `those banks' out of the subject clause, which goes against the rule.”
\end{quote}

\textit{Analysis:} Sonnet omits "those banks" from the sentential subject, misidentifying the full constituent as just "the patients' cleaning". As a result, it misapplies the constraint. This illustrates a case where the explanation is understood, but a failure in constituent recognition leads to incorrect reasoning.

\vspace{0.5em}
\textbf{(iii) GP-CoT(o1) prompt (GP success case)}\\
\textit{Excerpt from Sonnet’s response:}
\begin{quote}
\small
“Let's approach this step-by-step using the grammar explanation provided \ldots Here, `The patients' cleaning [of] those banks' is a complex subject that includes its own clause. It acts as a single unit\ldots. According to the explanation, when forming questions, we shouldn't move words from inside the complex subject clause. The clause acts like a protective bubble\ldots In Sentence B, the word order suggests that `those banks' has been moved out of the subject clause and placed at the end. This violates the rule of not moving elements from within the subject clause\ldots.”
\end{quote}

\textit{Analysis:} Sonnet correctly identifies the full sentential subject, understands the constraint (described in the grammar prompt as a ‘bubble’), and selects the sentence that violates it. The reasoning is structurally sound and consistent with the grammatical explanation.

\vspace{0.5em}
\textit{Conclusion:} In the CoT failure case, the model fails to supply appropriate categories of analysis, does not produce accurate rules, and relies on surface-level heuristics. In GP-CoT cases, the model reasons using categories and rules introduced in the prompt. GP failures arise when constituents are misidentified or reasoning chains are mismatched to the conclusion. In success cases, GP prompts enable correct identification of grammatical structure, appropriate constraint application, and accurate, rule-based reasoning. As models improve in reasoning, grammar prompting may become even more effective.

\subsection{Chinese Wh-fronting (GPT-3.5)}%
GPT-3.5 performs strongly on Chinese wh-fronting tasks in zero-shot settings (e.g., 98.3\% accuracy), but shows a noticeable drop under grammar prompting with chain-of-thought (GPb+CoT) conditions -- falling to 88.7\% (Table \ref{tab:SLING_main_table}). This reflects a broader issue: grammar prompting introduces an additional layer of explicit reasoning, which GPT-3.5 does not always handle reliably.

\textit{Wh-fronting}, common in English, involves moving a wh-word to the front of the sentence (e.g., “I will hear it” becomes “What will I hear?”). However, in Chinese, such fronting is ungrammatical. The correct structure keeps the wh-word in situ: “I will hear what?” Thus, illegal wh-fronting is easily detectable to native speakers, and GPT-3.5 reliably chooses the correct option under basic conditions.

When prompted to reason explicitly about this rule, GPT-3.5 sometimes falters. Consider the following example from the wh\_mod paradigm:

\begin{tcolorbox}[
colframe=black, colback=purple!10,
boxrule=0.8pt, sharp corners,
width=\columnwidth,
]
\begin{itemize}[leftmargin=1em, nosep]
\item[(A)] \textit{Tamen jiakuai} [VERB] \textit{le shenme} [WHAT] \textit{fazhan}? \checkmark
\item[(B)] \textit{Shenme} [WHAT] \textit{fazhan tamen jiakuai} [VERB] \textit{le}? $\times$
\end{itemize}
\end{tcolorbox}

\textit{Excerpt from GPT-3.5's response:}
\begin{quote}
“\ldots In Sentence B, the question word \textit{shenme} ("what") comes before the verb \ldots Based on the grammar explanation provided, Sentence B is more grammatically acceptable because the question word \textit{shenme} ("what") is correctly placed after the verb, following the rule that question words typically stay in the same position as the information they're asking about would be in a statement.”
\end{quote}

\textit{Analysis:} GPT-3.5 begins with a correct observation that the wh-word appears before the verb in Sentence B. However, it later contradicts itself, claiming that the wh-word is after the verb and choosing the wrong sentence. This suggests a failure in internal consistency: although it correctly tracks the constituent position at first, it loses track of its earlier analysis during its reasoning chain. Such cases reveal that GPT-3.5 struggles with maintaining reasoning context and may misapply rules it appears to have learned.

\vspace{0.5em}
\textit{Conclusion:} GPT-3.5 exhibits characteristic reasoning failures in GP-CoT settings. These include:
\begin{itemize}[nosep]
\item failure to track prior statements consistently,
\item misapplication of otherwise correctly identified grammatical rules,
\item distraction by irrelevant surface features.
\end{itemize}
While these issues are especially visible in GPT-3.5, similar patterns emerge across more complex paradigms even in stronger models. Grammar prompting clearly adds interpretive structure to a model’s linguistic reasoning process but that process itself remains susceptible to error.

\end{document}